\documentclass[10pt,twocolumn,letterpaper]{article}

\usepackage[pagenumbers]{cvpr} 

\usepackage[accsupp]{axessibility} 

\definecolor{cvprblue}{rgb}{0.21,0.49,0.74}
\usepackage[pagebackref,breaklinks,colorlinks,citecolor=cvprblue]{hyperref}
\setlist[itemize]{noitemsep, topsep=0pt}

\usepackage{hyperref}
\usepackage{url}
\usepackage{graphicx}
\usepackage{booktabs}
\usepackage{enumitem}
\usepackage{multirow}
\usepackage{wrapfig}
\usepackage{xcolor,colortbl}
\usepackage{subcaption}
\usepackage{amssymb}
\usepackage{pifont}
\def\rvc{{\mathbf{c}}}

\def\rvs{{\mathbf{s}}}

\def\rvx{{\mathbf{x}}}

\def\rvz{{\mathbf{z}}}

\def\paperID{16768} 
\def\confName{CVPR}
\def\confYear{2025}
\begin{document}
\title{SUGAR: Subject-Driven Video Customization in a Zero-Shot Manner}
\author{Yufan Zhou,~Ruiyi Zhang,~Jiuxiang Gu,~Nanxuan Zhao,~Jing Shi,~Tong Sun \\
Adobe Research \\
{\tt\small \{yufzhou, ruizhang, jigu, nanxuanz,  jingshi, tsun\}@adobe.com}
}
\maketitle
\begin{abstract}
    We present SUGAR, a zero-shot method for subject-driven video customization. Given an input image, SUGAR is capable of generating videos for the subject contained in the image and aligning the generation with arbitrary visual attributes such as style and motion specified by user-input text. Unlike previous methods, which require test-time fine-tuning or fail to generate text-aligned videos, SUGAR achieves superior results without the need for extra cost at test-time. To enable zero-shot capability, we introduce a scalable pipeline to construct synthetic dataset which is specifically designed for subject-driven customization, leading to 2.5 millions of image-video-text triplets. Additionally, we propose several methods to enhance our model, including special attention designs, improved training strategies, and a refined sampling algorithm. Extensive experiments are conducted. Compared to previous methods, SUGAR achieves state-of-the-art results in identity preservation, video dynamics, and video-text alignment for subject-driven video customization, demonstrating the effectiveness of our proposed method.
\end{abstract}    
\section{Introduction}
Subject-driven customization~\citep{ruiz2023dreambooth,chen2023suti,zhou2024CAFE,wu2023tune-a-video,wei2024dreamvideo,jiang2024videobooth} seeks to create images or videos for a specific identity which only appears in a user-provided image or video. The generated images and videos are also expected to align with arbitrary requirements specified in user-input text.
Although some efforts have been made in subject-driven video customization, existing approaches struggle to generate videos with good identity preservation and text-alignment.

In this work, we propose \textbf{SUGAR}, a novel method that performs \textbf{SU}bject-driven, customized video \textbf{G}ener\textbf{A}tion in a ze\textbf{R}o-shot manner. 
Compared to existing methods, our SUGAR can generate videos with better identity preservation and text-alignment. Specifically, SUGAR can extract identity-only information from user-input image, and generate videos for the target subject with a style, texture and motion guided by arbitrary user-input text. Some generated examples are shown in Figure \ref{fig:example}.

Our approach is inspired by some recent works on subject-driven image customization~\citep{chen2023suti,zhou2024CAFE,zhou2024toffee}. In this work, we explore two key questions: how to build a large-scale synthetic dataset specifically designed for subject-driven video customization to enable zero-shot capability, and how to enhance model performance through improved model design, training strategy, and sampling algorithm. Our major contributions can be summarized as follows:

\begin{figure*}[t!]
    \centering
    \includegraphics[width=0.9\linewidth]{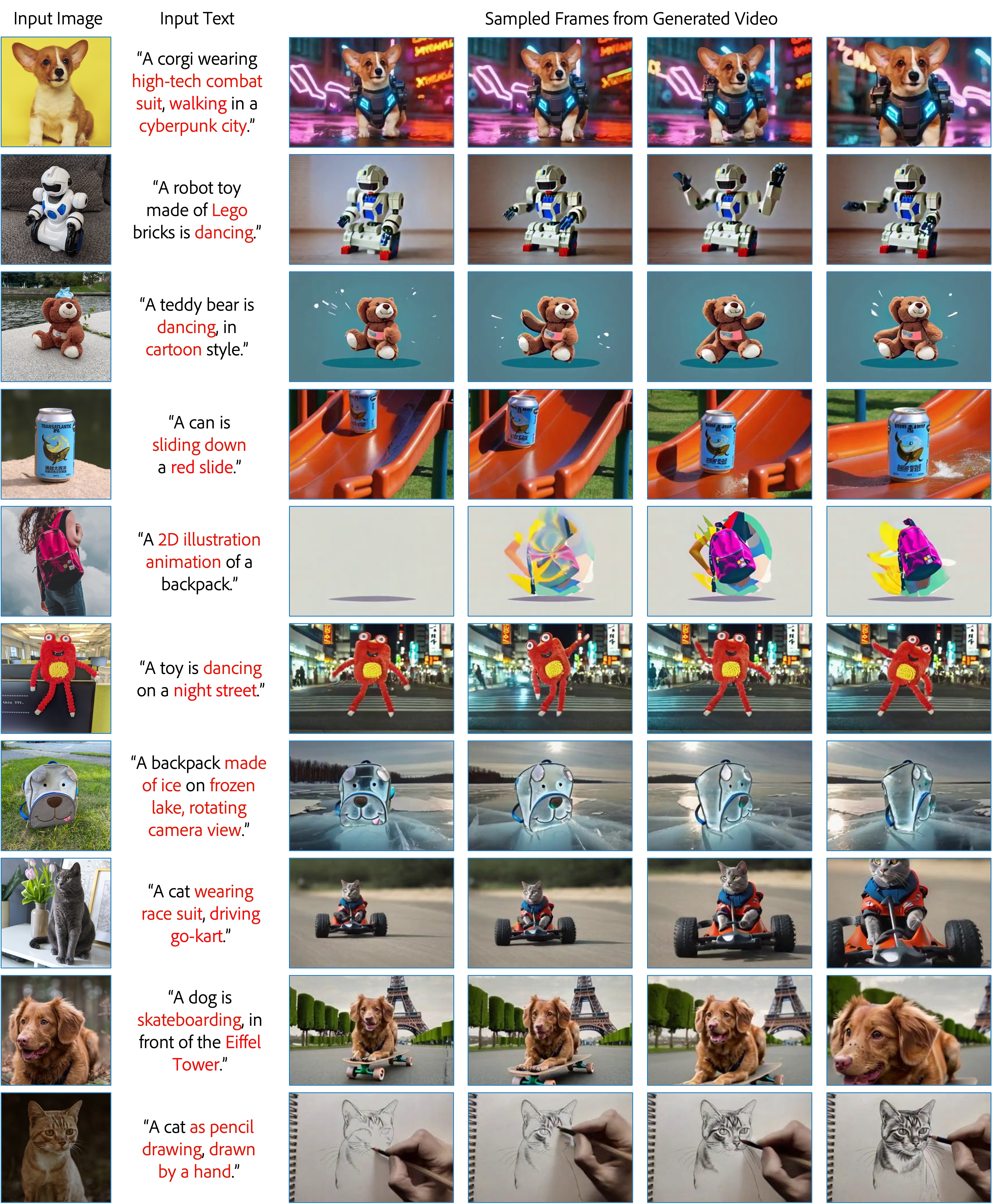}
    \vspace{-0.1in}
    \caption{Generated examples from the proposed method, where the frames are randomly sampled from the generated video. Our proposed method will generate videos for a specific subject contained in user-input image, in a zero-shot manner. The generated video will also meet the requirements described by user-input text.}
    \label{fig:example}
    \vspace{-0.25in}
\end{figure*}

 \begin{itemize}
     \item We propose a novel method for zero-shot subject-driven video customization. Given single subject image, our method generates high-quality video for the subject, while adhering to the requirements such as specified style or motion from input text;
     \item We propose a dataset construction pipeline and construct a dataset containing 2.5 million image-video-text triplets. As shown in our experiments, our synthetic dataset is necessary for achieving good text alignment;
     \item We propose and test various model designs, training strategies, and sampling algorithms, which significantly improve the performance of our final model;
     \item We conducted extensive experiments, and our proposed method outperforms all previous methods across different metrics. Some ablation studies are also conducted to provide readers with a deeper understanding of the proposed method;
 \end{itemize}
\section{Related Works}
Recently, researchers have achieved significant progress in text-guided image and video generation. Large-scale pre-trained models are capable of generating text-aligned images~\citep{ramesh2021zero, saharia2022imagen,rombach2022LDM, esser2024sd3,FLUX} or videos~\citep{Blattmann_2023_align_your_latents,singer-make-a-video,ho2022imagenvideo,kondratyuk2023videopoet,wang2023lavie,blattmann2023stablevideodiffusion,Sora,RunwayGen3,LumaDreamMachine,polyak2024moviegen,gupta2025photorealistic-video} with high-quality and diverse content.
However, despite of their impressive capabilities, these models fail to perform customized generation for novel concepts, such as generating creative images or videos for a specific subject from single user-provided testing image.

Various approaches have been proposed for the customized generation task in both image and video domains. Textual Inversion~\citep{gal2022textualinversion} proposes to represent the target with a text embedding which can be optimized inside the text embedding space of pre-trained text-to-image (T2I) generation model. DreamBooth~\citep{ruiz2023dreambooth} and CustomDiffusion~\citep{kumari2022multi} propose to fine-tune the pre-trained T2I model on the testing image so that fine-grained details of the subject can be captured. In the domain of text-to-video (T2V) generation, some methods~\citep{wu2023tune-a-video,wei2024dreamvideo} also propose to fine-tune pre-trained T2V models or adapters, on user-input subject image or video so that the resulting models can generate customized videos for the target subject. 

However, aforementioned approaches require extra time and computation cost due to the need of fine-tuning pre-trained models, which is inefficient for real-world applications. 
Thus some researchers investigate test-time tuning-free methods which can generate customized images or videos in a zero-shot manner, without the need for test-time fine-tuning~\citep{shi2023instantbooth,chen2023anydoor,pan2023kosmos,ma2023subjectdiffusion,zhou2024CAFE,chen2023suti,zhou2024toffee,polyak2024moviegen,wei2024dreamvideo2}. Our proposed SUGAR, inspired by these works, also aims to generate subject-driven customized videos with better quality in a zero-shot manner.

Different from a concurrent work DreamVideo-2~\citep{wei2024dreamvideo2}, which also focuses on zero-shot subject-driven T2V generation, we consider a transformer-based rather than a UNet-based diffusion model. Additionally, we introduce a novel pipeline for constructing synthetic datasets and explore distinct model design, training strategy and sampling algorithm, setting our work apart from DreamVideo-2. 

\section{Method}

\begin{figure}[t!]
    \centering
    \includegraphics[width=0.9\linewidth]{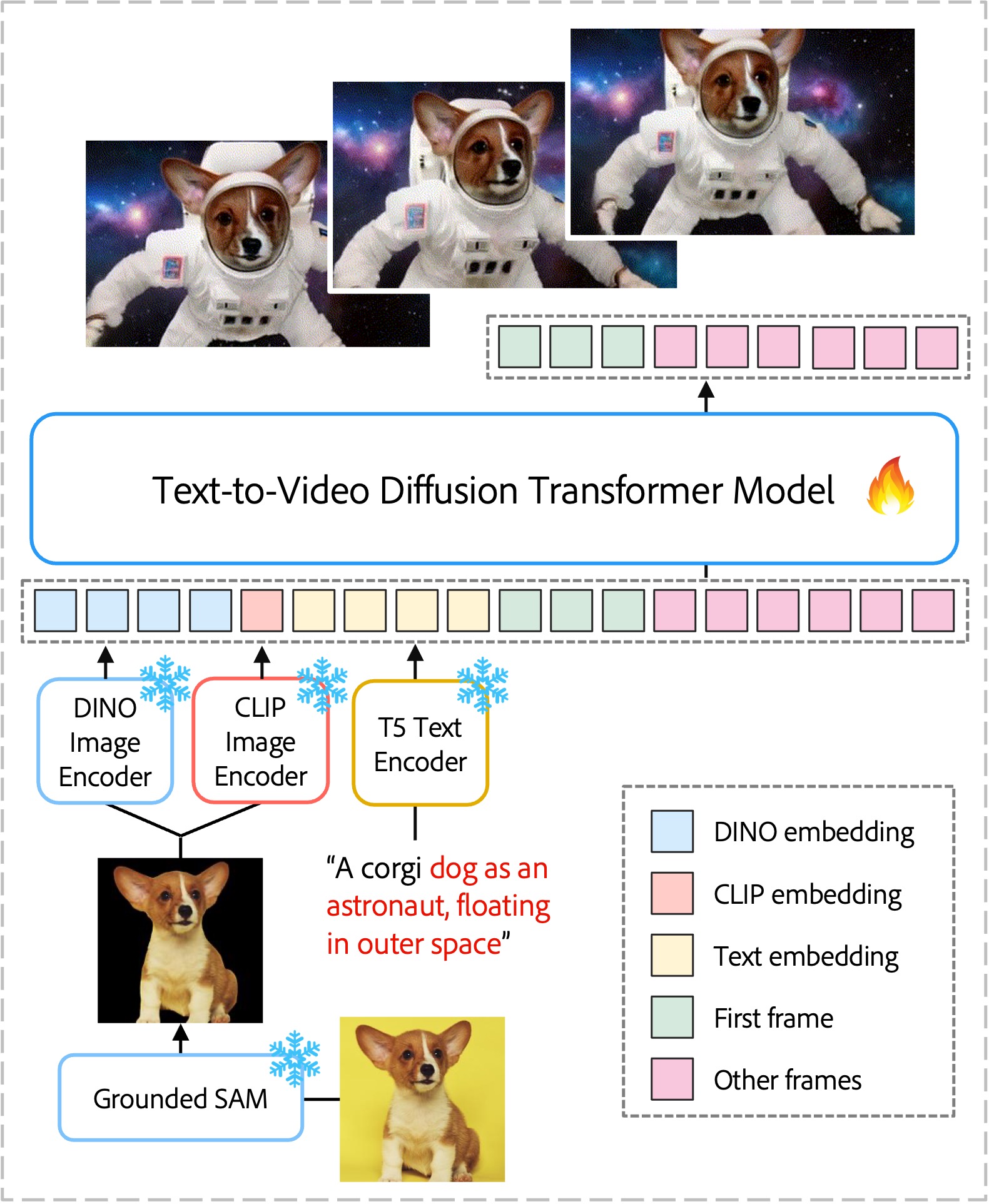}
    \vspace{-0.1in}
    \caption{Illustration of our model, randomly sampled frames from the generated video are shown in the figure for better illustration.}
    \label{fig:sugar_model}
    \vspace{-0.2in}
\end{figure}

Recent works~\cite{chen2023suti,zhou2024CAFE,zhou2024toffee} have shown that training generative models on high-quality synthetic dataset leads to impressive results in zero-shot image customization.
Inspired by these works, we start with constructing a synthetic dataset designed for subject-driven video customization. Then we  discuss how to better utilize this synthetic data with improved model design, training strategy and sampling algorithm.

Throughout the paper, we use $\rvs$ to represent the input subject image and use $\rvz$ to denote the identity image, which contains only the target subject without any background. The identity image $\rvz$ is obtained by applying segmentation on $\rvs$, using pre-trained Grounded SAM~\citep{ren2024GroundedSAM}.
We use $\rvc$ to denote user-input text specifying requirements such as video style and subject motion. The target video is represented by $\rvx$. Some pre-trained encoders are used in our approach, including pre-trained DINOv2~\citep{oquab2024dinov2} image encoder, CLIP~\citep{radford2021CLIP} image encoder and T5 text encoder, which are represented by $f_{\text{D}}(\cdot)$, $f_{\text{C}}(\cdot)$ and $f_{\text{T5}}(\cdot)$ respectively. 

Our model design is illustrated in Figure \ref{fig:sugar_model}, we design our method by extending the architecture of CogVideoX~\citep{yang2024cogvideox}, which is a transformer-based diffusion model defined in the latent space of a pre-trained Variational Auto-encoder~\citep{kingma2013auto}. The input of our model is the concatenation of DINO embedding, CLIP image embedding, text embedding and noisy latent code. Different projection layers are applied respectively before concatenation. 

\subsection{Synthetic Dataset Construction}
\begin{figure}[t!]
    \centering
    \includegraphics[width=0.99\linewidth]{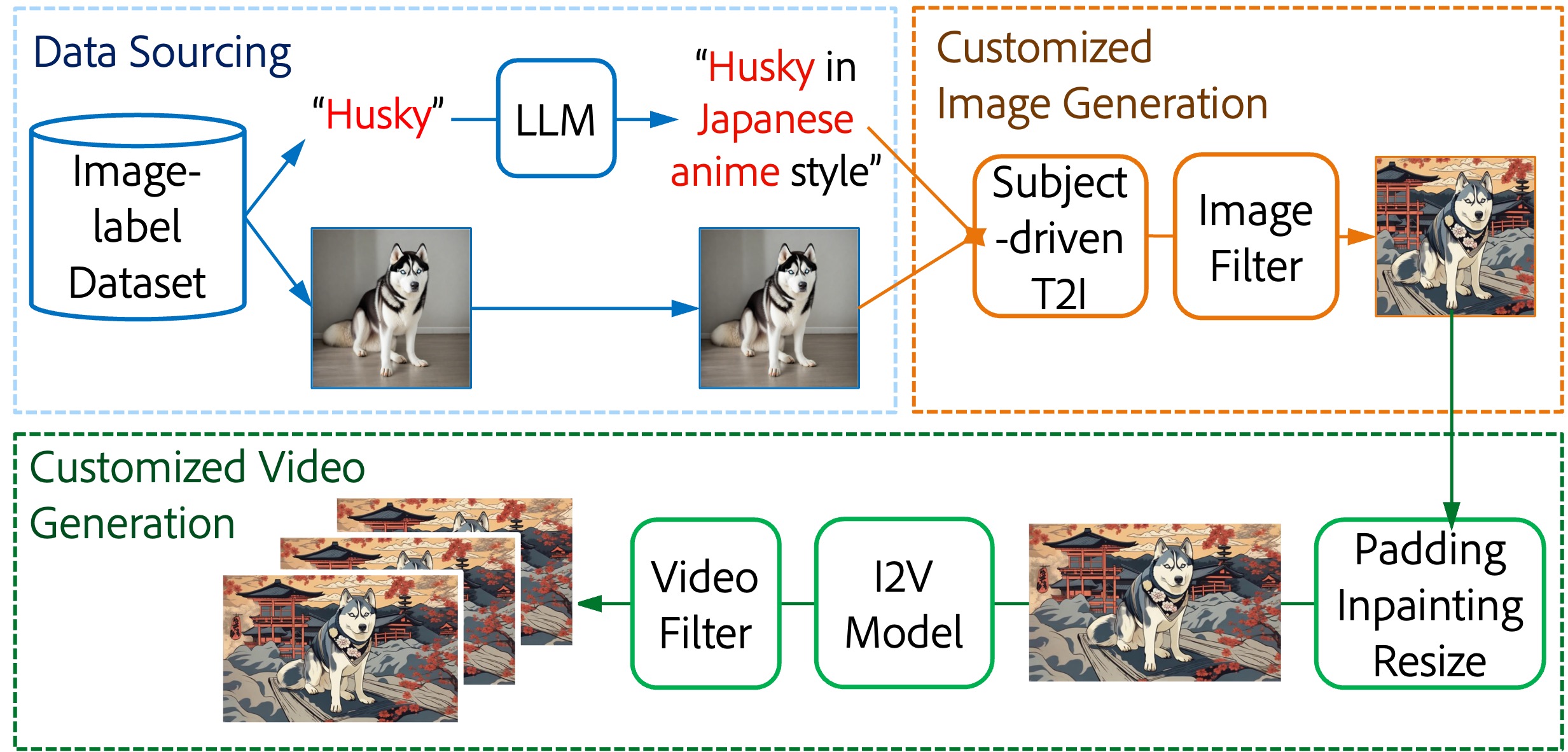}
    \vspace{-0.1in}
    \caption{The proposed pipeline for synthetic data generation.}
    \label{fig:dataset_construction}
    \vspace{-0.2in}
\end{figure}
Recall that given a subject image $\rvs$, our target is to generate a video $\rvx$, which should contain the desired identity and be aligned with user-input text $\rvc$. To obtain a model that can perform zero-shot video customization, we would like to train the model on a large-scale dataset consisting of triplet samples in the form of $(\rvx, \rvc, \rvs)$. Specifically, we expect $\rvx$ and $\rvs$ to contain the same identity but differ in terms of visual attributes such as style, color, etc., as required by text $\rvc$. Training with such a dataset leads to a model that extracts identity-only information from the input image, leading to more flexible and text-aligned generation.

Our synthetic dataset is generated with the pipeline presented in Figure \ref{fig:dataset_construction}, with details provided as follows. 

\paragraph{Data sourcing} 
We start with a collection of image-label samples featuring millions of subjects, covering a wide range of subjects including objects and animals. 
For each image, we generate a text prompt which simulates user intention, describing a target video containing the subject with specified visual attributes such as style, texture, color, or background.
Specifically, we have a pre-defined set of text templates which is first manually designed then enriched by large language model~\citep{touvron2023llama2}.

\paragraph{Customized image generation} 
Given an image-text pair from the previous step, we utilize the existing method to perform subject-driven text-to-image generation, resulting in a customized image based on the specified subject and text.
Specifically, we choose to follow Toffee~\citep{zhou2024toffee} for this step because of its superior performance and efficiency. 
We filter out the generated image when it is not aligned with the text, or the identity deviates too much from the input image. The image-text alignment and identity preservation are evaluated by cosine similarity between features extracted with pre-trained CLIP~\citep{radford2021CLIP} and DINO~\citep{caron2021dino}.

\paragraph{Customized video generation} 
Given an image customization result from the previous step, we perform data processing steps, including padding, inpainting, and resizing to modify its aspect ratio and resolution. The processed image is then fed into a pre-tained image-to-video (I2V) model to generate a dynamic video.
Due to computation and efficiency issues, we use Stable Diffusion XL inpainting model~\citep{podellsdxl} and DynamiCraft~\citep{xing2023dynamicrafter}. The dataset quality can be further improved by adopting better pre-trained models.
At last, we apply video-level filtering, to filter out low-quality videos. For instance, we evaluate subject consistency by computing average DINO~\citep{caron2021dino} similarity between consecutive frames, and similarity between randomly sampled frames, to filter out videos with inconsistent subjects. Inspired by previous work~\citep{huang2023VBench}, we also filter out static videos using optical flow~\citep{teed2020raft};

In the end, we obtain a dataset containing 2.5 millions of image-video-text triplet samples.
In later experiments, we will show that this synthetic dataset plays an important role and hugely improves the performance of resulting model.

\subsection{Model Training and Testing}
Different from subject-driven T2I generation~\citep{ruiz2023dreambooth,gal2022textualinversion,kumari2023multi,shi2023instantbooth,wei2023elite,zhou2023profusion,ma2023subjectdiffusion,chen2023suti,zhou2024CAFE,zhou2024toffee} which focus on identity preservation and text-alignment, subject-driven video generation further takes subject motion into consideration. 
However, we find that most samples from our synthetic dataset are videos of standing still subjects. This is because the pre-trained I2V generation model we used often fails to generate text-aligned subject motions. 
As a result, our model directly trained with the constructed dataset also fails to generate text-aligned subject motion. To solve this problem, we propose the followings.

\paragraph{Utilizing Real-world Video}
To improve the subject motion of generated videos, we proposed to include a large-scale video-text dataset in our training. Large-scale real-world datasets~\citep{chen2024Panda-70M,bain2021WebVid} often contain videos with a variety of subject motions which we would like our model to learn. 
Different from our dataset, the real-world video-text dataset does not have the input image. To bridge this gap, we use Grounded SAM~\cite{ren2024GroundedSAM} and perform object segmentation on randomly sampled frames to obtain the identity image. 

A limitation of real-world dataset is that it is difficult to find frames within the same video which contain the same subject with different styles or textures. As a result, identity images and video frames from real-world videos will share the same style, subject texture, and other visual attributes. Training a model on such data makes it challenging to generate video in styles that is different from the input image.
In our experiments, we find that utilizing both real-world and synthetic datasets leads to better performance than training the model on only synthetic or real-world datasets. 

\paragraph{Training Strategy}
A consequent question is how to train the model with both synthetic and real-world datasets. 
Throughout the paper, we assume that training samples are sampled from both real-world and synthetic datasets. We use $p$ to denote the probability of sampling from the synthetic dataset. As a result, $p=0$ stands for training the model only on real-world dataset, while $p=1$ meaning training solely on synthetic dataset. 
In this work, we propose the following training strategies:
\begin{itemize}
    \item SUGAR-Mix: A simple training strategy where we train the model on a mixed dataset with fixed $p$ during training. We try different $p$ selected from $\left[0.0, 0.5, 1.0 \right]$ in our experiments;
    \item SUGAR-TS: A two-stage training strategy, where TS stands for two-stage. The model is first trained on real-world videos, which corresponds to $p=0.0$. Then the model is trained on mixed dataset with $p=0.5$;
    \item SUGAR-TSF: A two-stage training strategy, where TSF stands for two-stage with frozen layers. The model is trained on real-world videos and mixed dataset sequentially. Different from SUGAR-TS, we freeze some layers in SUGAR-TSF during the second stage training. Our assumption is that the layers closer to the input side determine the basic semantic and motion of the generated video. By freezing these layers in second stage, the model retains its motion generation capability learned from real-world videos, while learning improved visual attribute change from synthetic videos. For simplicity, we keep the first half of the model layers frozen in our experiments;
\end{itemize}

\paragraph{Special Attention Design}
To mitigate the effects of learning limited motions from synthetic video dataset, we propose some special designs for attention operation in the transformer.
Our assumption is that not all the frames have to be directly conditioned on the input image embeddings. Although image embeddings are important for maintaining subject identity, they also contain rich structure information that leads to generated videos with little subject motion. Our new attention mechanism is termed as Selective Attention, highlighting that embeddings can only attend to specific selected others.

\begin{figure}[t!]
    \centering
    \includegraphics[width=0.85\linewidth]{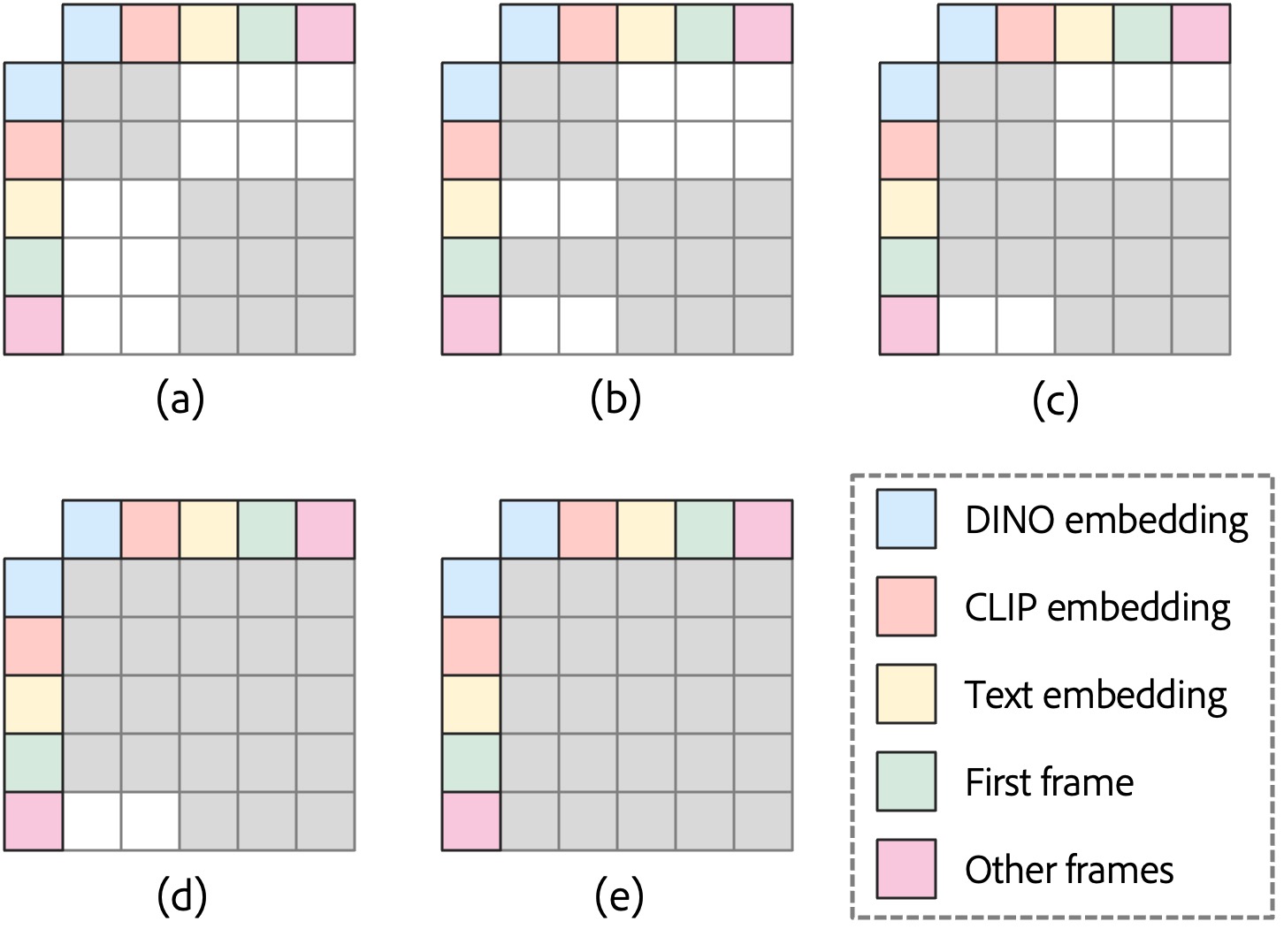}
    \vspace{-0.1in}
    \caption{Different attention designs of our proposed model. One embedding can attend to another one only when the corresponding position is marked shadow in the above illustration. For instance, in design (b) image embeddings can not attend to the first frame, but the first frame can attend to the image embeddings.}
    \label{fig:attn_mask}
    \vspace{-0.1in}
\end{figure}

We propose several special designs as illustrated in Figure \ref{fig:attn_mask}, which presents how different embeddings attend to each other. 
For the ease of illustration, we simply use length of 1 for all kinds of embeddings in the figure. It is important to note that design (a) is equivalent to a T2V model which can not perform customized generation, as its video output is not conditioned on the input image. Meanwhile, design (e) is a simple extension of pre-trained T2V generation model, making it a baseline design for comparison.
In a later section, we will show how our proposed attention design can improve the subject motion of generated videos without the loss of identity preservation.

\paragraph{Improved Sampling}
Classifier-free guidance (CFG)~\citep{ho2021cfg} has been shown to play an important role in sampling from diffusion models. To enable CFG sampling, we drop DINO, CLIP, and text embeddings with a probability of 0.5, 0.2, and 0.2 independently during training.
Instead of using the same level of guidance for all conditions,
we follow \cite{brooks2023instructpix2pix} to use different levels of guidance for identity and text conditions, which provides better flexibility in generation. 
We propose two variants to modify score estimation:
\begin{align}\label{eq:two_condition_1}
     & \tilde{\epsilon}_{\theta} = \epsilon_\theta (\rvx_t, \emptyset_{\text{D}}, \emptyset_{\text{C}}, \emptyset_{\text{T}} )  \\
     & + \omega_I (\epsilon_\theta (\rvx_t, f_{\text{D}}(\rvz), f_{\text{C}}(\rvz), f_{\text{T5}}(\rvc))
     - \epsilon_\theta (\rvx_t,\emptyset_{\text{D}}, \emptyset_{\text{C}},  f_{\text{T5}}(\rvc)) \nonumber \\
      & + \omega_T (\epsilon_\theta (\rvx_t, \emptyset_{\text{D}}, \emptyset_{\text{C}}, f_{\text{T5}}(\rvc))
     - \epsilon_\theta (\rvx_t, \emptyset_{\text{D}}, \emptyset_{\text{C}}, \emptyset_{\text{T}})),\nonumber
\end{align}
and
\begin{align}\label{eq:two_condition_2}
     & \tilde{\epsilon}_{\theta} = \epsilon_\theta (\rvx_t, \emptyset_{\text{D}}, \emptyset_{\text{C}}, \emptyset_{\text{T}} )  \\
     & + \omega_T (\epsilon_\theta (\rvx_t, f_{\text{D}}(\rvz), f_{\text{C}}(\rvz), f_{\text{T5}}(\rvc))
     - \epsilon_\theta (\rvx_t, f_{\text{D}}(\rvz), f_{\text{C}}(\rvz),  \emptyset_{\text{T}}) \nonumber \\
      & + \omega_I (\epsilon_\theta (\rvx_t, f_{\text{D}}(\rvz), f_{\text{C}}(\rvz),  \emptyset_{\text{T}})
     - \epsilon_\theta (\rvx_t, \emptyset_{\text{D}}, \emptyset_{\text{C}}, \emptyset_{\text{T}})), \nonumber
\end{align}
where $\emptyset_{\text{D}}, \emptyset_{\text{C}}$ represent embeddings for unconditional case, $ \emptyset_{\text{T}}$ denotes the text embedding of negative text prompt, $\omega_I$ and $\omega_T$ denote the guidance scales for image and text respectively~\cite{ho2021cfg, brooks2023instructpix2pix}.

To further improve the subject motion at inference time, we propose to drop image embeddings at the early stage of sampling. Dropping image embeddings provides more flexibility, leading to generated videos with better text-alignment and subject motion. However, dropping image embeddings also leads to identity loss as we will show in the experiments. To solve this problem, we propose to only drop $f_{\text{D}}(\rvz)$ and keep using $f_{\text{C}}(\rvz)$ which contains coarse information of the input identity. As we will show in experiments, this approach keeps the generated identity closely aligned with the target while achieving improved subject motion. 

Let $0 \leq \bar{t}\leq T$ be a hyper-parameter, for timestep $\bar{t} \leq t \leq T$, we revise the estimation \eqref{eq:two_condition_1} and \eqref{eq:two_condition_2} to be
\begin{align}\label{eq:two_condition_3}
     & \tilde{\epsilon}_{\theta} = \epsilon_\theta (\rvx_t, \emptyset_{\text{D}}, \emptyset_{\text{C}}, \emptyset_{\text{T}} )  \\
     & + \omega_I (\epsilon_\theta (\rvx_t, \emptyset_{\text{D}}, f_{\text{C}}(\rvz), f_{\text{T5}}(\rvc))
     - \epsilon_\theta (\rvx_t,\emptyset_{\text{D}}, \emptyset_{\text{C}},  f_{\text{T5}}(\rvc)) \nonumber \\
      & + \omega_T (\epsilon_\theta (\rvx_t, \emptyset_{\text{D}}, \emptyset_{\text{C}}, f_{\text{T5}}(\rvc))
     - \epsilon_\theta (\rvx_t, \emptyset_{\text{D}}, \emptyset_{\text{C}}, \emptyset_{\text{T}})), \nonumber
\end{align}
and
\begin{align}\label{eq:two_condition_4}
     & \tilde{\epsilon}_{\theta} = \epsilon_\theta (\rvx_t, \emptyset_{\text{D}}, \emptyset_{\text{C}}, \emptyset_{\text{T}} )  \\
     & + \omega_T (\epsilon_\theta (\rvx_t, \emptyset_{\text{D}}, f_{\text{C}}(\rvz), f_{\text{T5}}(\rvc))
     - \epsilon_\theta (\rvx_t, \emptyset_{\text{D}}, f_{\text{C}}(\rvz),  \emptyset_{\text{T}}) \nonumber \\
      & + \omega_I (\epsilon_\theta (\rvx_t, \emptyset_{\text{D}}, f_{\text{C}}(\rvz),  \emptyset_{\text{T}})
     - \epsilon_\theta (\rvx_t, \emptyset_{\text{D}}, \emptyset_{\text{C}}, \emptyset_{\text{T}})). \nonumber
\end{align}
In our experiments, \eqref{eq:two_condition_2} and \eqref{eq:two_condition_4} works slightly better than \eqref{eq:two_condition_1} and \eqref{eq:two_condition_3}.

\section{Experiment}

\subsection{Evaluation Metric}
Following previous benchmarks on subject-driven image customization~\citep{ruiz2023dreambooth} and text-to-video generation~\citep{huang2023VBench}, we propose to evaluate our model with the following metrics, which stand for some requirements we care about the most in subject-driven video customization.

\begin{table*}[t!]
    \centering
    \scalebox{0.9}{
    \begin{tabular}{lcccccc}
    \toprule
        \multirow{2}{*}{\textbf{Method}} & \textbf{DINO}  & \textbf{CLIP} & \textbf{ViCLIP} & \textbf{Dynamic} & \textbf{Subject}  & \textbf{Background} \\
         & \textbf{Score} ($\uparrow$) & \textbf{Score} ($\uparrow$) & \textbf{Score} ($\uparrow$) & \textbf{Degree} ($\uparrow$)& \textbf{Consis.} ($\uparrow$) & \textbf{Consis.} ($\uparrow$) \\
        \midrule
        \textbf{VideoBooth} & 0.493 & 0.266 & 0.173 & 0.512 & 0.938 & 0.960  \\
        \textbf{DreamVideo} & 0.376 & 0.296 & 0.213 & 0.722 & 0.875 & 0.937  \\
        \textbf{Vidu-1.5} & 0.654 & 0.303 & 0.237 & 0.645 & 0.928 & 0.942 \\
        \midrule
        \textbf{Subject-driven T2I + I2V} &\\
         ~~~~~~ + DynamiCrafter & 0.660 & 0.321 & 0.260 & 0.420 & 0.970 & 0.971 \\
         ~~~~~~ + CogVideoX & 0.642 & 0.323 & 0.264 & 0.325 & 0.969 & 0.967 \\
         ~~~~~~ + Runway Gen-3 & 0.650 & 0.326 & 0.265 & 0.522 & 0.944 & 0.950 \\
         ~~~~~~ + Luma Dream Machine & 0.578 & 0.320 & 0.270 & 0.634 & 0.915 & 0.935 \\
        \midrule
        \textbf{SUGAR (Ours)} & \\
         ~~~~~~  $\omega_T$=7.5, $\omega_I$=7.5 & \cellcolor{blue!10} \underline{0.742}  & 0.311 & 0.250 & 0.608 &  \cellcolor{blue!10} \underline{0.979} &  \cellcolor{blue!10} \underline{0.983}\\
         ~~~~~~ $\omega_T$=5.0, $\omega_I$=7.5 &  \cellcolor{blue!10} 0.723 & 0.318 & 0.262 &  0.687 &  \cellcolor{blue!10} 0.976 &  \cellcolor{blue!10} 0.981 \\
         ~~~~~~  $\omega_T$=4.0, $\omega_I$=7.5 &  \cellcolor{blue!10} 0.708 & 0.322 & 0.267 &  \cellcolor{blue!10} 0.737 &  \cellcolor{blue!10} 0.972 &  \cellcolor{blue!10} 0.980\\
         ~~~~~~ $\omega_T$=3.0, $\omega_I$=7.5 &  \cellcolor{blue!10} 0.684 & \cellcolor{blue!10} 0.327 & \cellcolor{blue!10} 0.273 &  \cellcolor{blue!10}0.757 &  \cellcolor{blue!10} 0.970 &  \cellcolor{blue!10} 0.977 \\
         ~~~~~~ $\omega_T$=2.5, $\omega_I$=7.5 &  \cellcolor{blue!10} 0.664 &  \cellcolor{blue!10} \underline{0.329} &  \cellcolor{blue!10} \underline{0.275} &  \cellcolor{blue!10} \underline{0.772} & 0.962 &  \cellcolor{blue!10} 0.974 \\
        \bottomrule
    \end{tabular}
    }
    \vspace{-0.1in}
    \caption{Quantitative evaluation of different methods. $\omega_T$ and $\omega_I$ are guidance scales for text and identity conditions, respectively. We use underline to indicate the best result. We also \colorbox{blue!10}{highlight} cells for our method when the results outperform those of all baseline methods.}
    \label{tab:main_results}
    \vspace{-0.2in}
\end{table*}

\paragraph{Identity preservation}
We use DINO~\citep{caron2021dino} score to assess whether the generated video contains the same subject as input image. Specifically, for a given video, we calculate the average cosine similarity between DINO features of the input image and each video frame. Intuitively, a higher DINO score reflects better identity preservation.

\paragraph{Text-alignment}
To assess how well the generated video aligns with input text, we calculate the feature similarity between video and text using pre-trained CLIP~\citep{radford2021CLIP} and ViCLIP~\citep{internvid_viclip}. Specifically, we compute the average CLIP similarity between all video frames and the text. Higher CLIP and ViCLIP scores indicate stronger text alignment.

\paragraph{Dynamic Degree}
Because we aim to generate a dynamic video rather than a static one, it is essential to evaluate whether a video contains large motions. Following ~\cite{huang2023VBench}, we use RAFT~\citep{teed2020raft} to estimate the dynamic degree of the generated video, with a higher score indicating better motion dynamics.

\paragraph{Consistency}
We expect the subject and background to remain consistent throughout the video. 
Following VBench~\citep{huang2023VBench}, we utilize pre-trained DINO and CLIP encoder to measure subject consistency and background consistency respectively. The detailed computation process can be found in \cite{huang2023VBench}, higher score indicates better consistency.


\subsection{Main Results}

We compare our method with baselines which fall into two categories: some approaches generate customized videos with single, end-to-end model, while others involve a sequential process. In the latter, we combine subject-driven image customization with image-to-video (I2V) generation methods, allowing us to obtain customized images and videos sequentially. 

We select
VideoBooth~\citep{jiang2024videobooth} and DreamVideo~\citep{wei2024dreamvideo}
as single-model baselines,
as they are well-known methods for video customization.
We also include Vidu-1.5~\citep{bao2024vidu,vidu-1.5} in comparison, which recently started supporting subject-driven video customization and made the service publicly accessible.
A concurrent work DreamVideo-2~\citep{wei2024dreamvideo2} requires an additional input besides subject image, which is the trajectory of subject bounding box, thus can not be directly compared with ours. 
For the sequential generation baseline, we choose Toffee~\citep{zhou2024toffee} for subject-driven image customization because of its strong performance, and utilize several popular or state-of-the-art I2V generation models including DynamiCrafter~\cite{xing2023dynamicrafter}, CogVideoX-5B~\citep{yang2024cogvideox}, Luma Dream Machine~\citep{LumaDreamMachine}, Runway Gen-3\footnote{We use Gen-3 Alpha Turbo as it is efficiently accessible via API.}~\citep{RunwayGen3}.

Following previous works~\citep{ruiz2023dreambooth,wei2024dreamvideo}, we use subject images provided by DreamBench~\citep{ruiz2023dreambooth} for evaluation. Specifically, DreamBench consists of 30 subjects including various animals and objects. We selected one image for each subject as shown in the Appendix. Different from DreamBench which is designed for subject-driven image customization, we would like to include motion requirements in our testing prompts. 
However, certain motions are only applicable to some subjects. 
For example, the motion ``dance" is suitable for a dog but not for a vase.
To address this, we categorize the subjects into 3 classes: animals, active objects which may perform human-like motions, and static objects. It is important to note that static objects can also exhibit motions such as sliding and floating.
We carefully designed 18 prompts for each category, which are provided in the appendix, resulting in 540 subject-prompt combinations. 

Our main quantitative results are shown in Table \ref{tab:main_results}.
Our SUGAR model is trained with attention design (b), training strategy SUGAR-TSF, and evaluated with different sampling hyper-parameters. We choose $\bar{t}=T=1000$ for simplicity unless otherwise specified.
From the results, we can find that our proposed method outperforms the previous method on different metrics across different settings. 
We observe that Vidu-1.5 demonstrates impressive identity preservation and video quality. 
However, it faces challenges when the input text aims to modify subject attributes like style, texture, or color. It either generates a video without target attribute changes or a video with the desired changes but suffers from identity loss, which compromises its performance.
Some qualitative results and discussions are provided in the Appendix.

\subsection{Ablation Study}
\paragraph{Comparison of different attention designs}
To understand the impact of different attention designs, we train several models with different designs under the same training strategy.
Specifically, we choose the simplest training strategy SUGAR-Mix with $p=0.5$, to eliminate the influence of training strategy and focus solely on the model design. 

\begin{figure}[t!]
    \centering
    \includegraphics[width=0.95\linewidth]{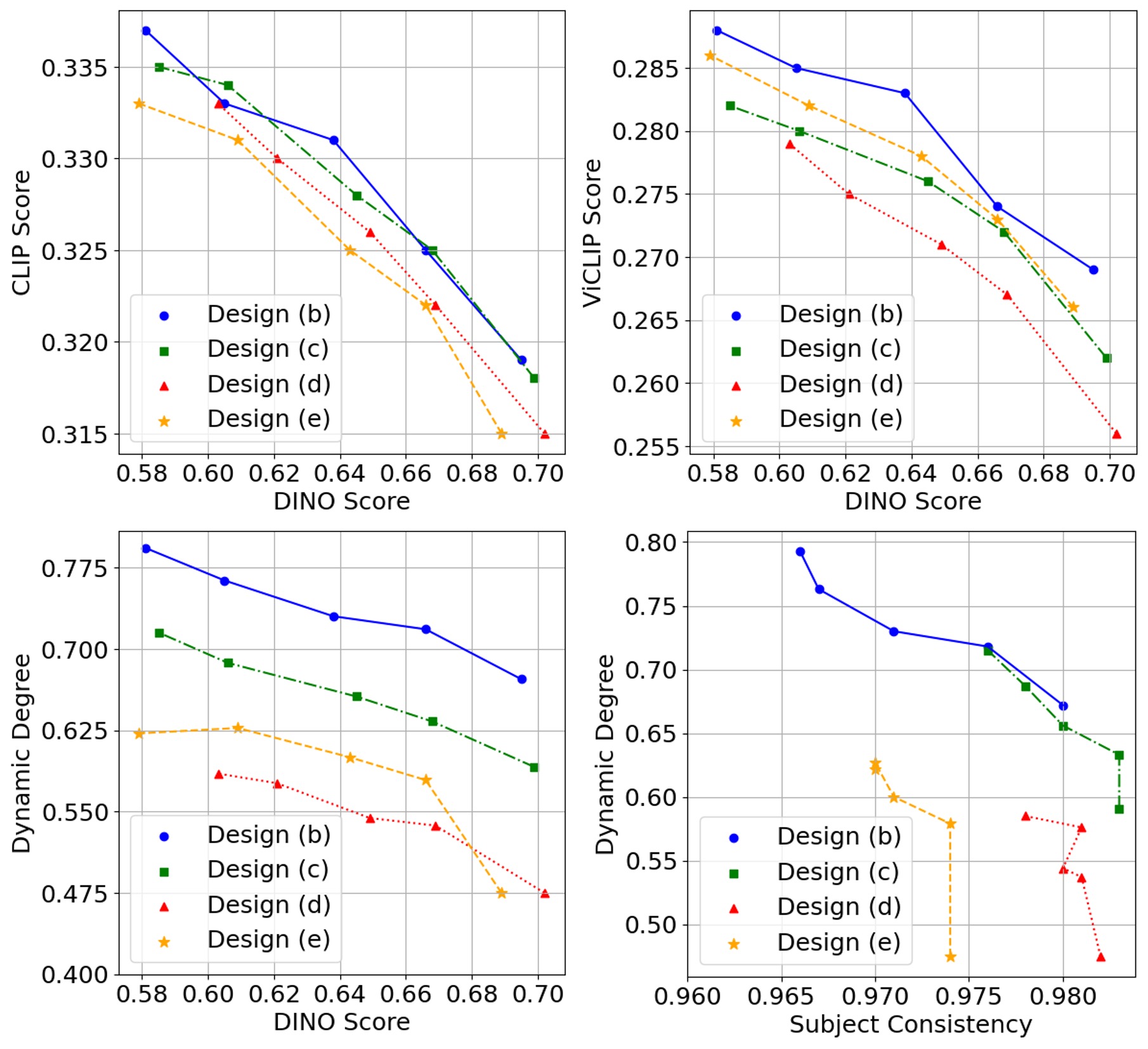}
    \vspace{-0.15in}
    \caption{Comparison of models with different attention designs.}
    \label{fig:ablation_attn}
    \vspace{-0.25in}
\end{figure}

The results are shown in Figure \ref{fig:ablation_attn}, where the models are evaluated with a collection of inference settings: we keep $\omega_T=7.5$ fixed and test the models with $\omega_I$ selected from $\{2.5, 3.0, 4.0, 5.0, 7.5\}$. We also test a model trained with attention design (a), whose results will not be influenced by different $\omega_I$ as the generation is not conditioned on the input image because of the attention design. It is not included in the figure because it only obtains a DINO score of 0.339. 
From the results, we can conclude that design (b) outperforms others in general. Even if only a single frame is directly conditioned on the input image, we do not observe subject identity loss. Furthermore, under the same level of identity preservation evaluated by the DINO score, design (b) obtains better text alignment as indicated by CLIP and ViCLIP scores, it also generates better motions, as indicated by a higher dynamic degree. 

\paragraph{Comparison of different training strategies}
To compare different training strategies, we train multiple models using different strategies while keeping the same attention design. In these experiments, we use the simplest attention design, which is design (e) from Figure \ref{fig:attn_mask}.

\begin{figure}[t!]
    \centering
    \includegraphics[width=0.95\linewidth]{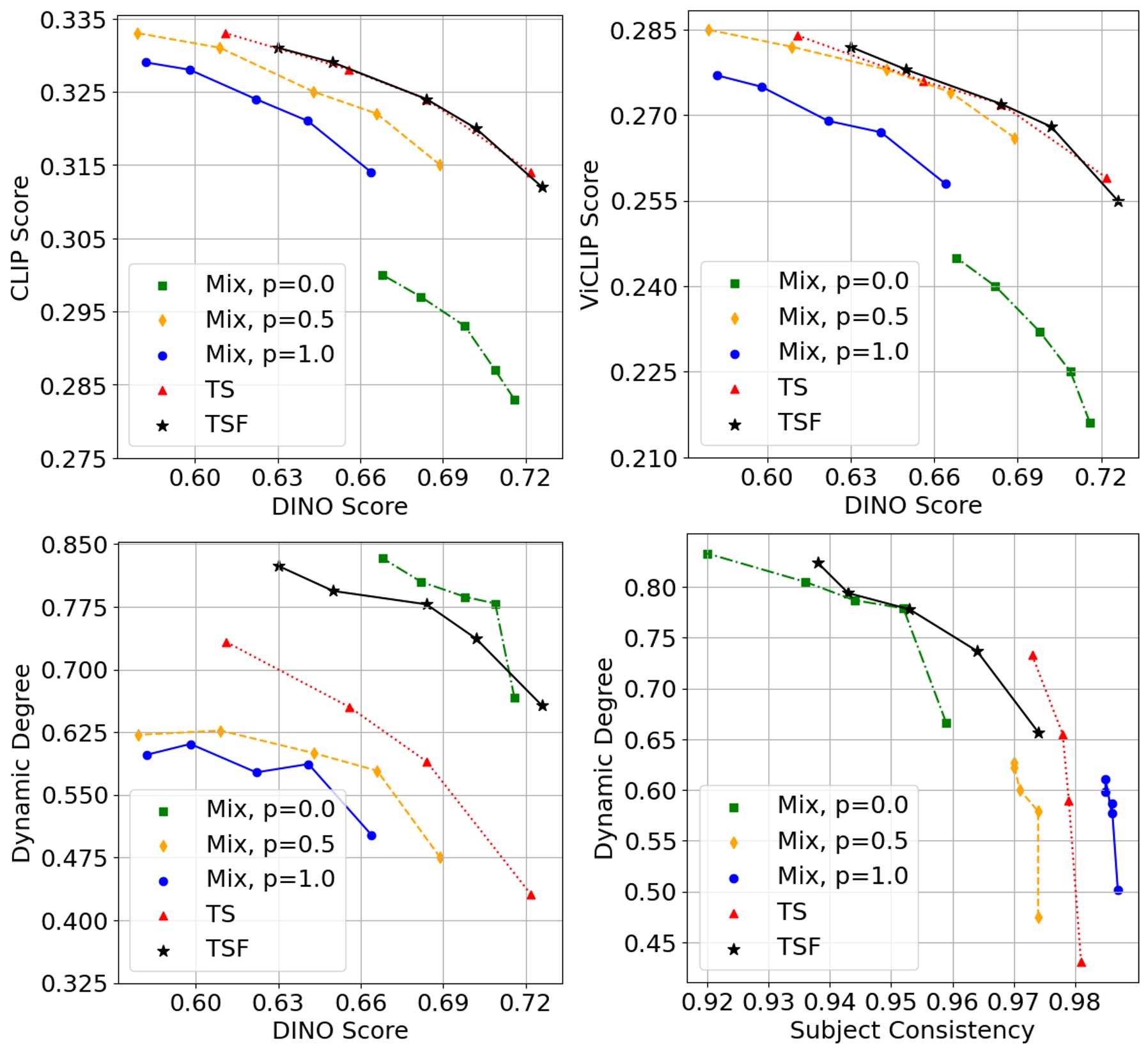}
    \vspace{-0.15in}
    \caption{Comparison of models using different training strategies.}
    \label{fig:ablation_training}
    \vspace{-0.25in}
\end{figure}

The results are shown in Figure \ref{fig:ablation_training}.
We notice that SUGAR-Mix with $p=0$ achieves competitive dynamic degree and much worse text-alignment compared to others. This is because the model is trained on samples processed from real-world videos, which provide rich information on subject motions but lack the style and color changes between the target videos and identity images.
On the other hand, SUGAR-Mix with $p=0.5$ or $p=1.0$ achieves better text-alignment due to the contribution of synthetic dataset, while the dynamic degree drops significantly. 
Among all the training strategies, our proposed SUGAR-TSF obtains good text-alignment without the loss of dynamic degree, showing the effectiveness of the proposed strategy.

\paragraph{The necessity of video customization dataset} 
In previous ablation study, we have already shown that the synthetic customization dataset is important to obtain good text-alignment. One remaining question is: instead of using the video customization dataset, what if we use the image customization dataset by treating the images as videos with the single frame? This is an important question as image customization dataset requires less construction cost, and many existing text-to-video generation models can be trained on videos and image datasets simultaneously. 

\begin{figure}[t!]
    \centering
    \includegraphics[width=0.95\linewidth]{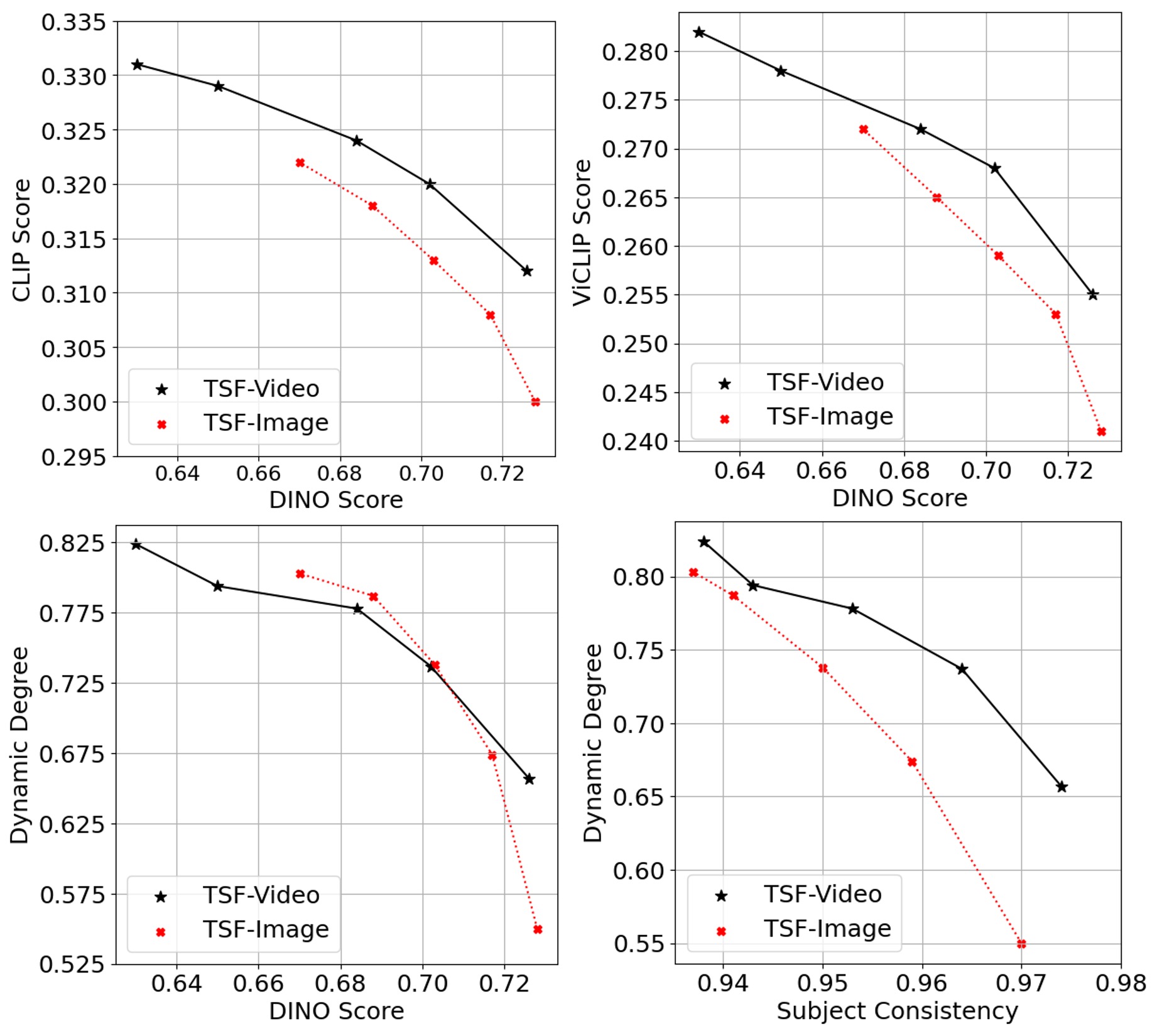}
    \vspace{-0.15in}
    \caption{Comparison of models trained with video customization dataset and image customization dataset.}
    \label{fig:ablation_vid_img}
    \vspace{-0.1in}
\end{figure}

We conduct an ablation study by training two models, TSF-Video and TSF-Image. Both models are trained with the SUGAR-TSF strategy and attention design (e).
We first train a model on real-world videos, then fine-tune it on two mixed datasets. TSF-Video is fine-tuned on a mixture of real-world videos and our synthetic dataset with $p=0.5$, TSF-Image is fine-tuned on a mixture of real-world videos and image customization dataset. For a fair comparison, we directly sample random frames from the video customization dataset as target images for TSF-Image during training.
The results are presented in Figure \ref{fig:ablation_vid_img}, from which we can see that TSF-Video obtains better results in text-alignment. TSF-Video also achieves a higher dynamic degree given the same level of subject consistency. We can conclude that video customization dataset provides stronger supervision for training than image customization dataset.

\begin{figure}[t!]
    \centering
    \includegraphics[width=0.9\linewidth]{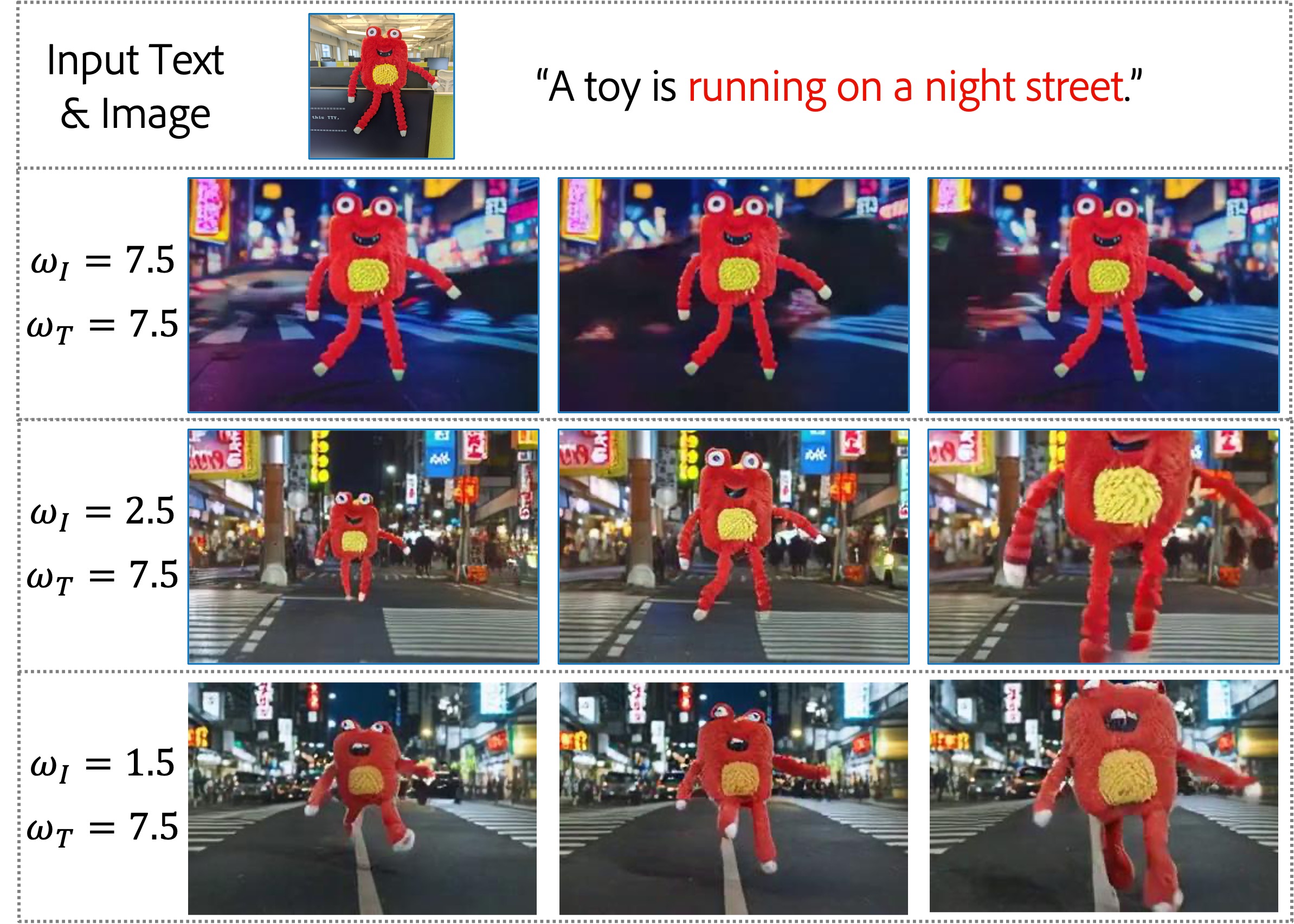}
    \vspace{-0.1in}
    \caption{Generated examples with different levels of guidance.}
    \label{fig:ablation_dual_condition}
    \vspace{-0.2in}
\end{figure}

\paragraph{Impact of dual conditions}
In addition to quantitative results of sampling using different $\omega_I, \omega_T$, we present qualitative comparison in Figure \ref{fig:ablation_dual_condition}, where we set $\bar{t}=T=1000$ to exclude the effect of dropping image embeddings. From the comparison, we can see that using large guidance for identity conditions leads to video with restricted motion, which can be improved by reducing $\omega_I$. However, when $\omega_I$ is too small, it may result in identity loss.

\paragraph{Impact of dropping image embedding}
We have shown that we can control the level of identity preservation and dynamic degree by using different $\omega_I$ and $\omega_T$. We would like to investigate the impact of dropping image embeddings at inference. Specifically, we set $\bar{t}=900$ and test 4 settings including dropping only DINO or CLIP embedding, dropping both embeddings and a model trained without dropping. 

\begin{figure}[t!]
    \centering
    \includegraphics[width=0.95\linewidth]{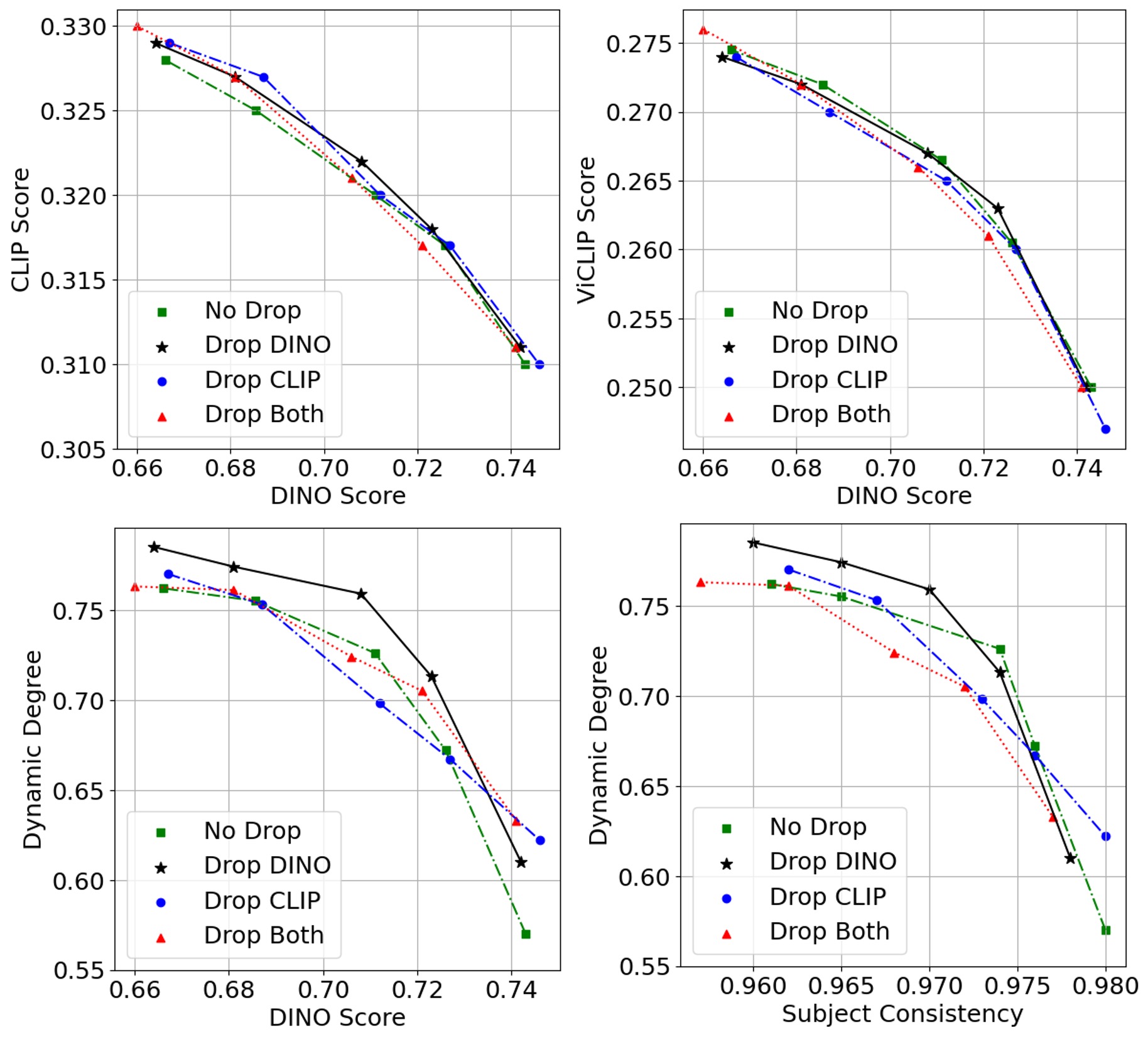}
    \vspace{-0.15in}
    \caption{Comparison of dropping different image embeddings.}
    \label{fig:ablation_drop}
    \vspace{-0.1in}
\end{figure}

\begin{figure}[t!]
    \centering
    \includegraphics[width=0.9\linewidth]{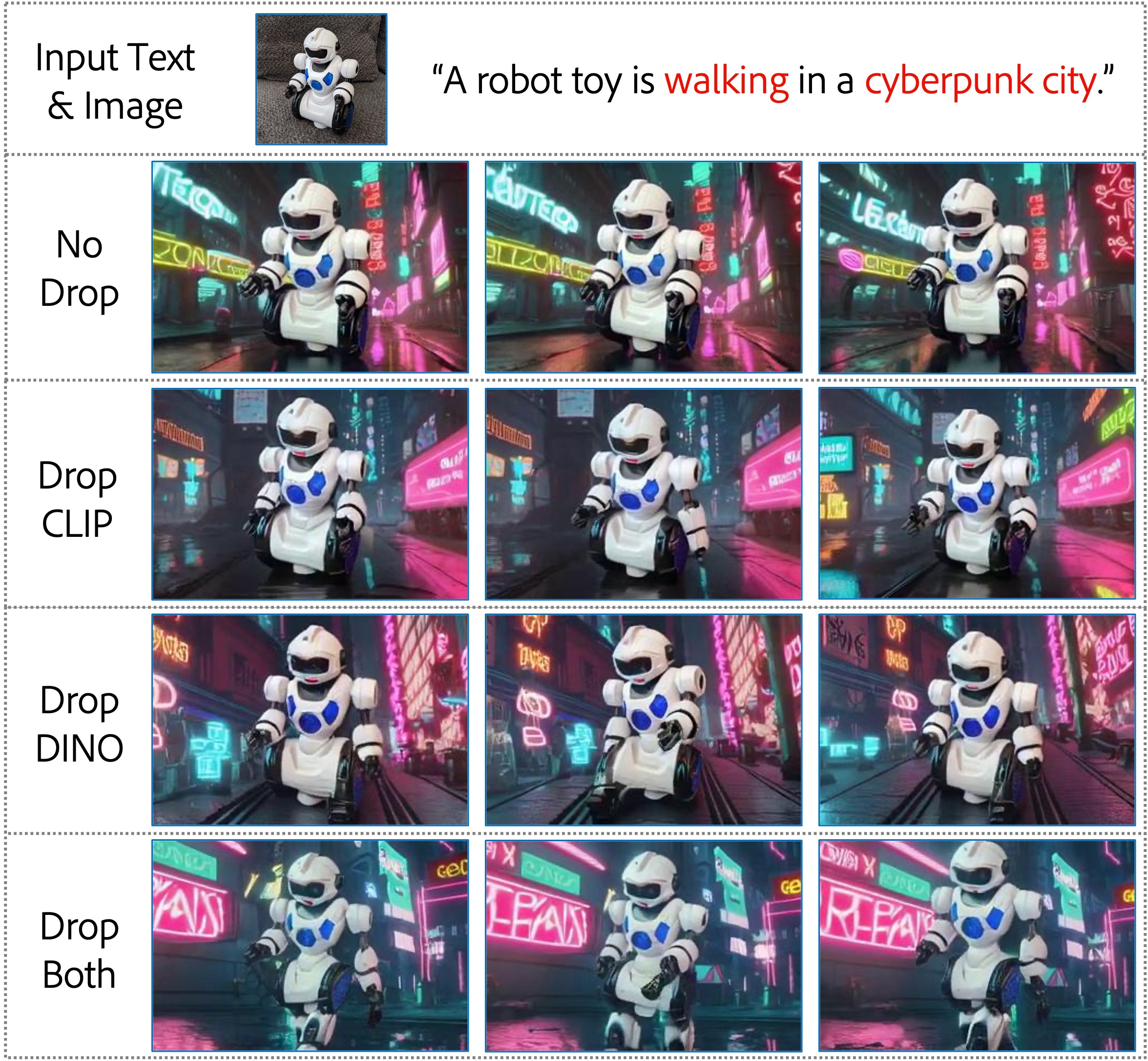}
    \vspace{-0.1in}
    \caption{Generated examples of dropping different embeddings.}
    \label{fig:ablation_drop_comparison} 
    \vspace{-0.2in}
\end{figure}

The results are shown in Figure \ref{fig:ablation_drop}, from which we can find that dropping DINO embedding at the early sampling stage leads to slightly better dynamic degree, without the loss of text-alignment and identity preservation. A qualitative comparison is shown in Figure \ref{fig:ablation_drop_comparison}, where we use $\omega_I=\omega_T=7.5$ to exclude the influence of dual condition sampling. From the comparison we can find that dropping image embedding indeed leads to better subject motion. However, dropping both DINO and CLIP embeddings may lead to identity loss: the robot in the figure has extra legs which do not exist in the input image.

\section{Conclusion}
We propose SUGAR, a zero-shot method for subject-driven video customization, without the need of test-time fine-tuning. Our approach consists of a dataset construction pipeline, a special attention design, a training strategy and improved sampling techniques. Extensive experiments demonstrate the effectiveness of our method, achieving state-of-the-art results.
\clearpage
{
    \small
    \bibliographystyle{ieeenat_fullname}
    \bibliography{main}
}
\clearpage
\appendix
\section{More Results}

\begin{figure}[t!]
    \centering
    \includegraphics[width=0.99\linewidth]{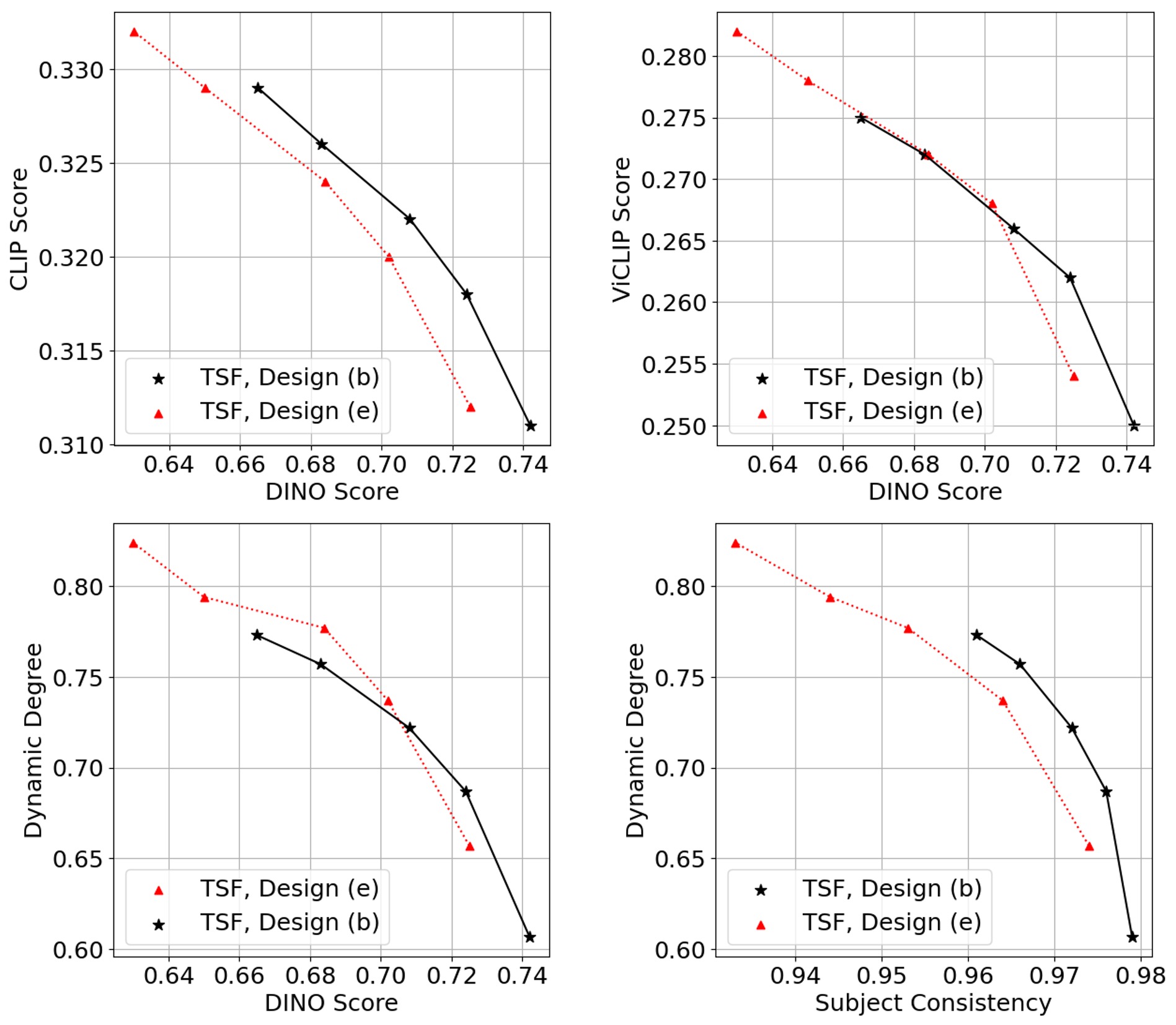}
    \vspace{-0.1in}
    \caption{Comparison of models with different attention designs.}
    \label{fig:appendix_ablation_attn}
    \vspace{-0.25in}
\end{figure}

\paragraph{More comparison of different attention designs} 
We have tested different attention designs where the models are trained with Sugar-Mix ($p=0.5$), and found that attention design (b) leads to the best results. Will this still hold true if we use our improved training strategy Sugar-TSF? To answer this question, we conduct another experiment whose results are shown in Figure \ref{fig:appendix_ablation_attn}. For clearer presentation, we only report the results of design (b) and design (e) here.
From the results, we can conclude that design (b) is still slightly better, while the performance gap becomes smaller, due to the contribution of our improved strategy Sugar-TSF.

\begin{figure}[t!]
    \centering
    \includegraphics[width=0.95\linewidth]{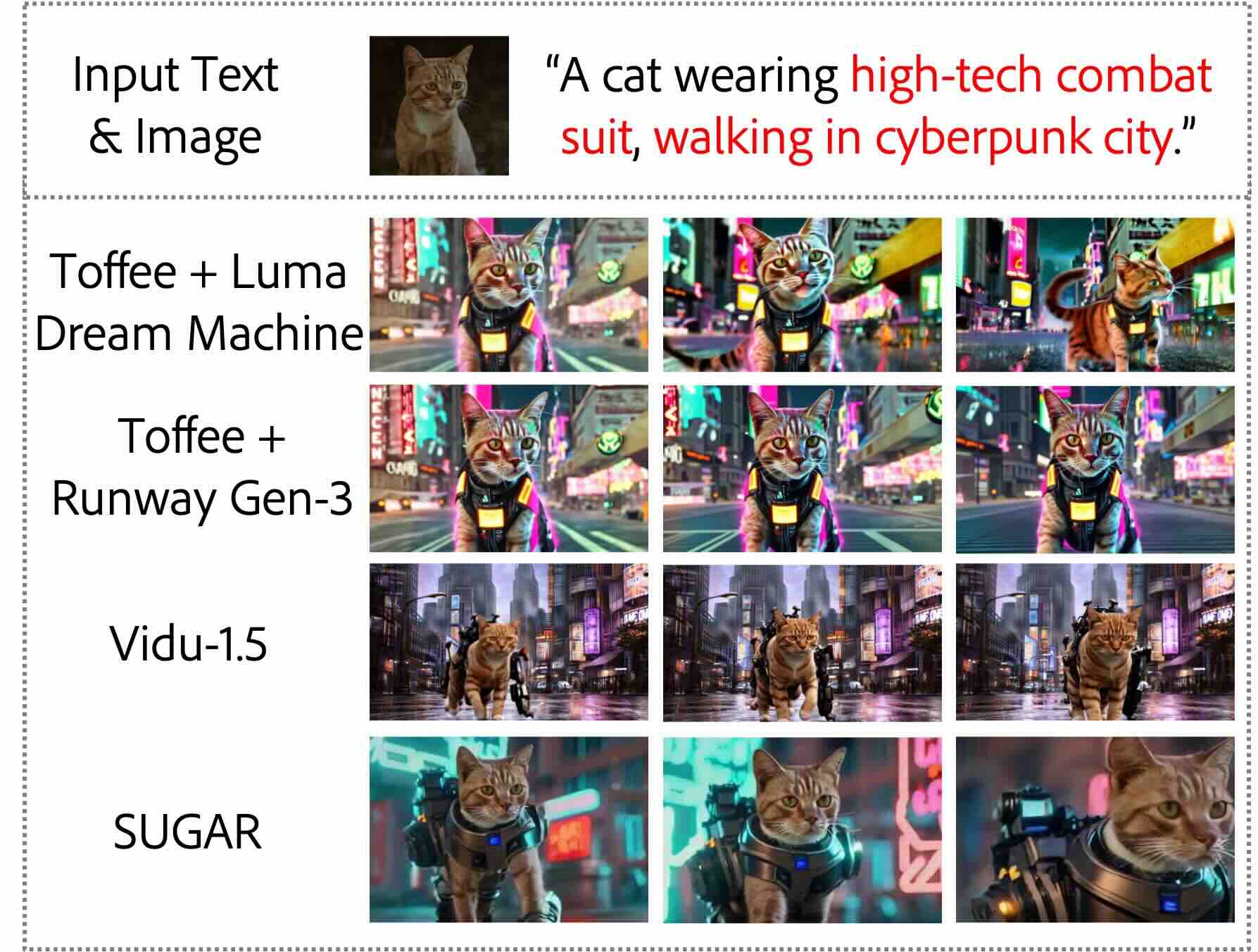}
    \vspace{-0.1in}
    \caption{Generated examples from SUGAR and baselines.}
    \label{fig:comparison_1}
    \vspace{-0.15in}
\end{figure}

\begin{figure}[t!]
    \centering
    \includegraphics[width=0.95\linewidth]{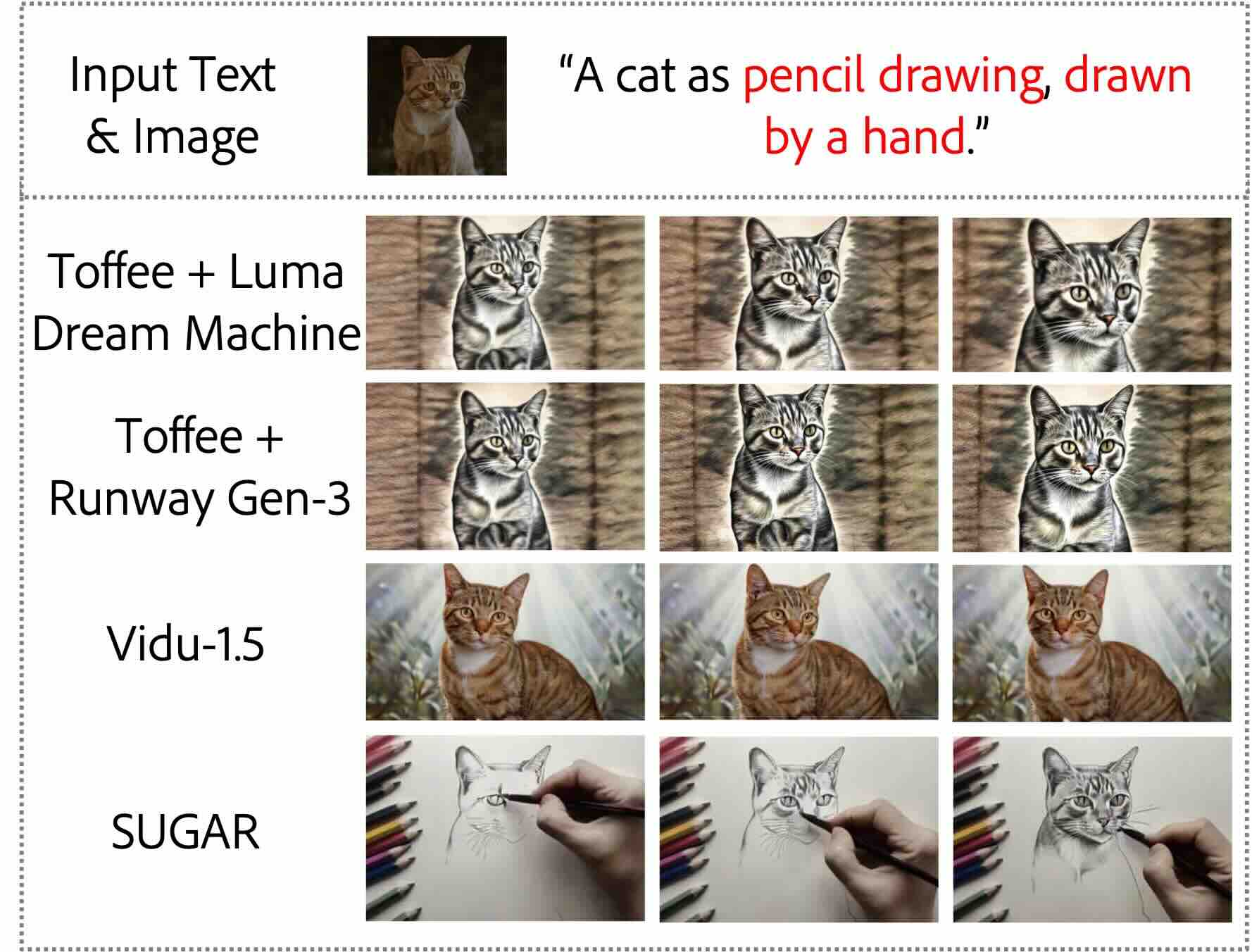}
    \vspace{-0.1in}
    \caption{Generated examples from SUGAR and baselines.}
    \label{fig:comparison_2}
    \vspace{-0.15in}
\end{figure}

\begin{figure}[t!]
    \centering
    \includegraphics[width=0.95\linewidth]{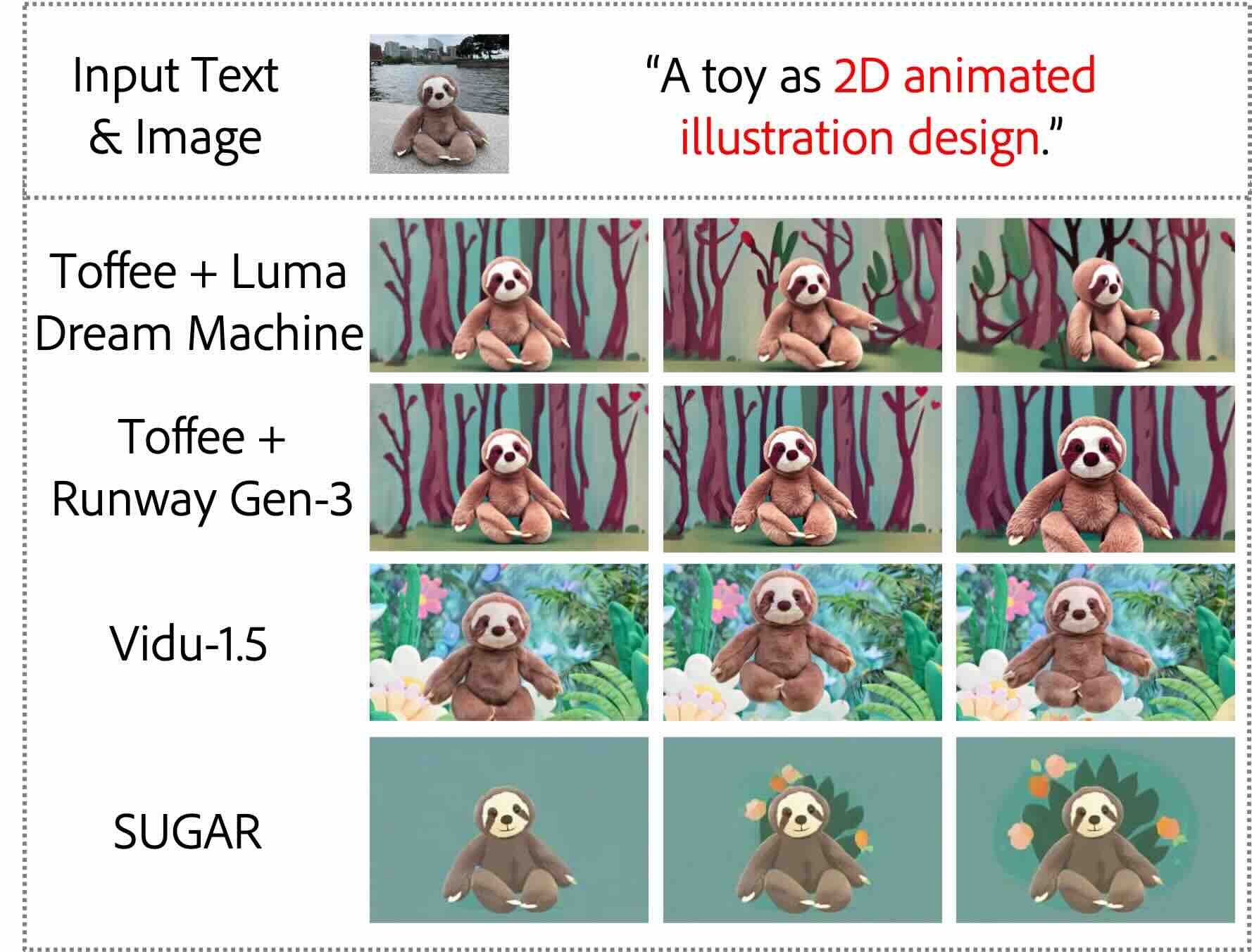}
    \vspace{-0.1in}
    \caption{Generated examples from SUGAR and baselines.}
    \label{fig:comparison_3}
    \vspace{-0.15in}
\end{figure}

\begin{figure}[t!]
    \centering
    \includegraphics[width=0.95\linewidth]{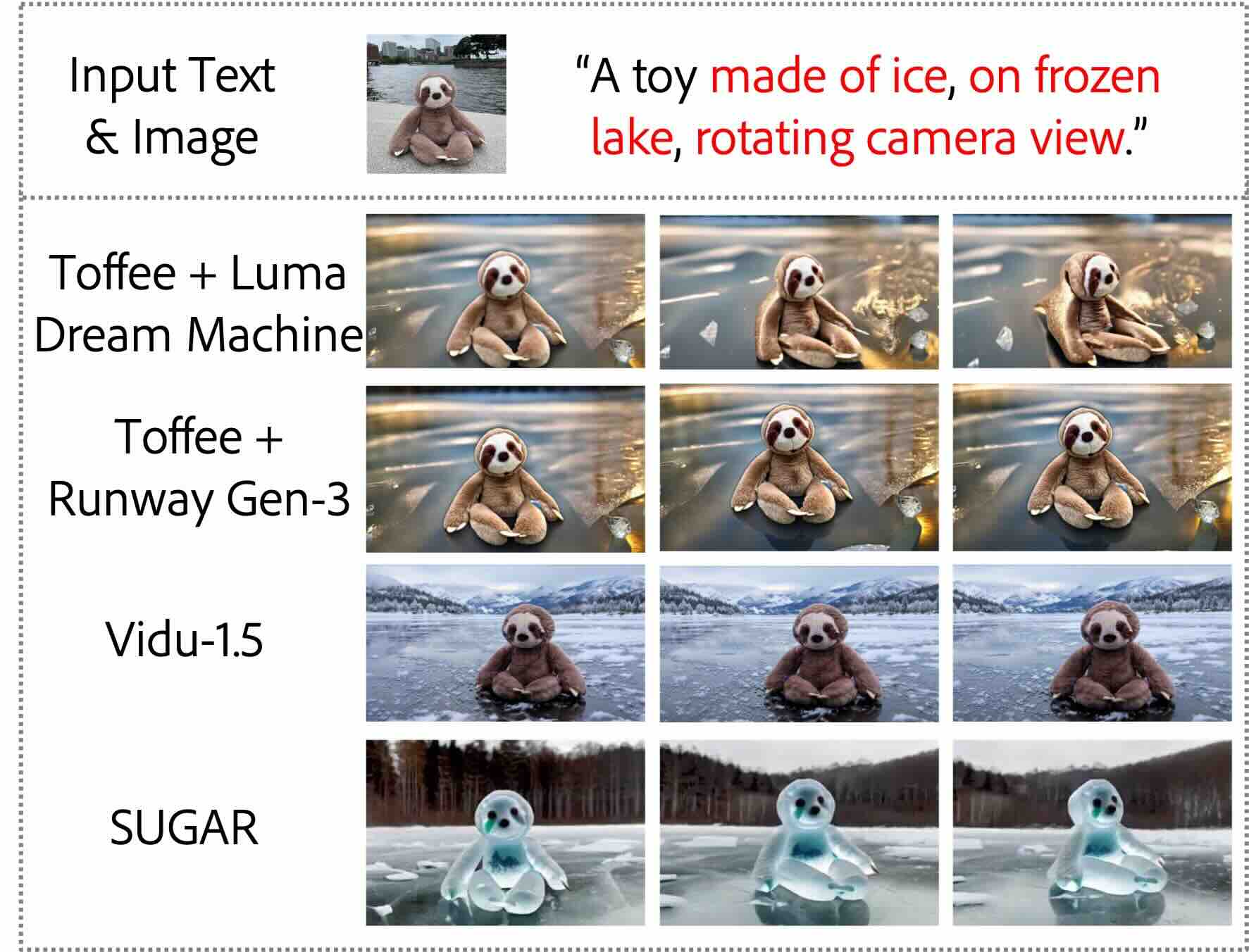}
    \vspace{-0.1in}
    \caption{Generated examples from SUGAR and baseline.}
    \label{fig:comparison_4}
    \vspace{-0.15in}
\end{figure}

\begin{figure}[t!]
    \centering
    \includegraphics[width=0.95\linewidth]{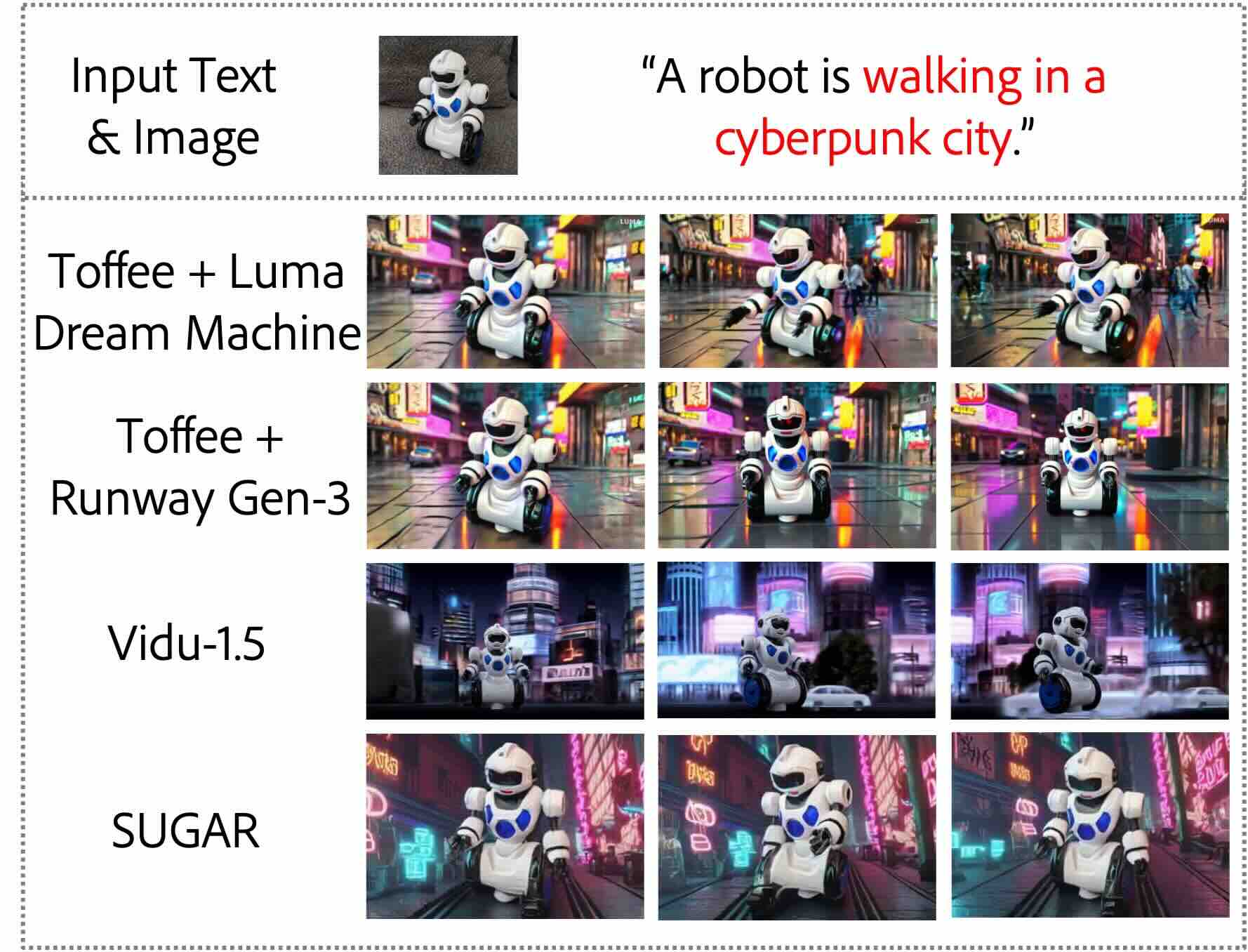}
    \vspace{-0.1in}
    \caption{Generated examples from SUGAR and baseline.}
    \label{fig:comparison_5}
    \vspace{-0.15in}
\end{figure}

\begin{figure}[t!]
    \centering
    \includegraphics[width=0.95\linewidth]{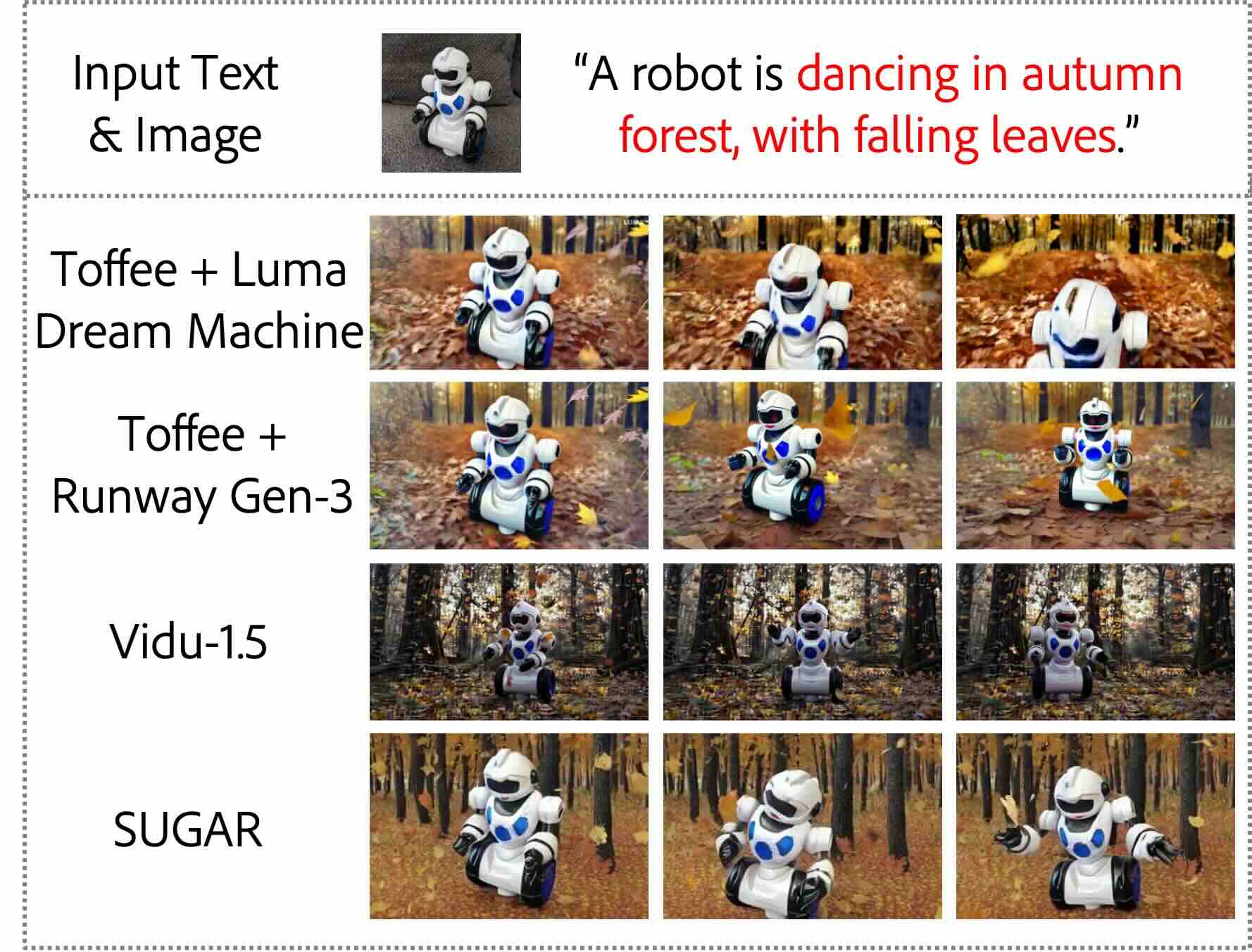}
    \vspace{-0.1in}
    \caption{Generated examples from SUGAR and baseline.}
    \label{fig:comparison_6}
    \vspace{-0.15in}
\end{figure}

\paragraph{More comparison with baseline methods}
We present some generated examples in Figure \ref{fig:comparison_1}, \ref{fig:comparison_2}, \ref{fig:comparison_3}, \ref{fig:comparison_4}, \ref{fig:comparison_5}, \ref{fig:comparison_6}. 

Baseline methods using sequential generation, which combine Toffee~\citep{zhou2024toffee} with Luma Dream Machine~\citep{LumaDreamMachine} and Runway Gen-3~\citep{RunwayGen3}, sometimes fail to maintain the subject identity.  This is because the pre-trained I2V models used in these methods are not trained on customization dataset, they may generate creative but undesired outputs. For example, the appearance of cat from Luma Dream Machine in Figure \ref{fig:comparison_1} changes; the toy from Luma Dream Machine in Figure \ref{fig:comparison_4} has a deformation issue. Another issue with sequential generation baseline is error propagation: if the subject-driven image customization fails to generate text-aligned image, it will cause the generated video to fail as well: in Figure \ref{fig:comparison_4} baselines fail to generate subject with texture change because the image customization fails.

Vidu-1.5~\citep{vidu-1.5}, although obtains good performance in some cases, fails to change the style or texture of the subject as shown in Figure \ref{fig:comparison_2}, \ref{fig:comparison_3}, \ref{fig:comparison_4}.

All the baseline methods face challenges in understanding concepts such as ``pencil drawing drawn by a hand" and ``2D animated illustration" in Figure \ref{fig:comparison_2} and \ref{fig:comparison_3}. They also fail to
generate certain subject motions. For instance, in Figure \ref{fig:comparison_5} and \ref{fig:comparison_6}, the robots in their generated videos seem to be ``sliding" rather than ``walking" or ``dancing" as required by the input text.

\section{More Details}
\begin{figure}[t!]
    \centering
    \includegraphics[width=0.99\linewidth]{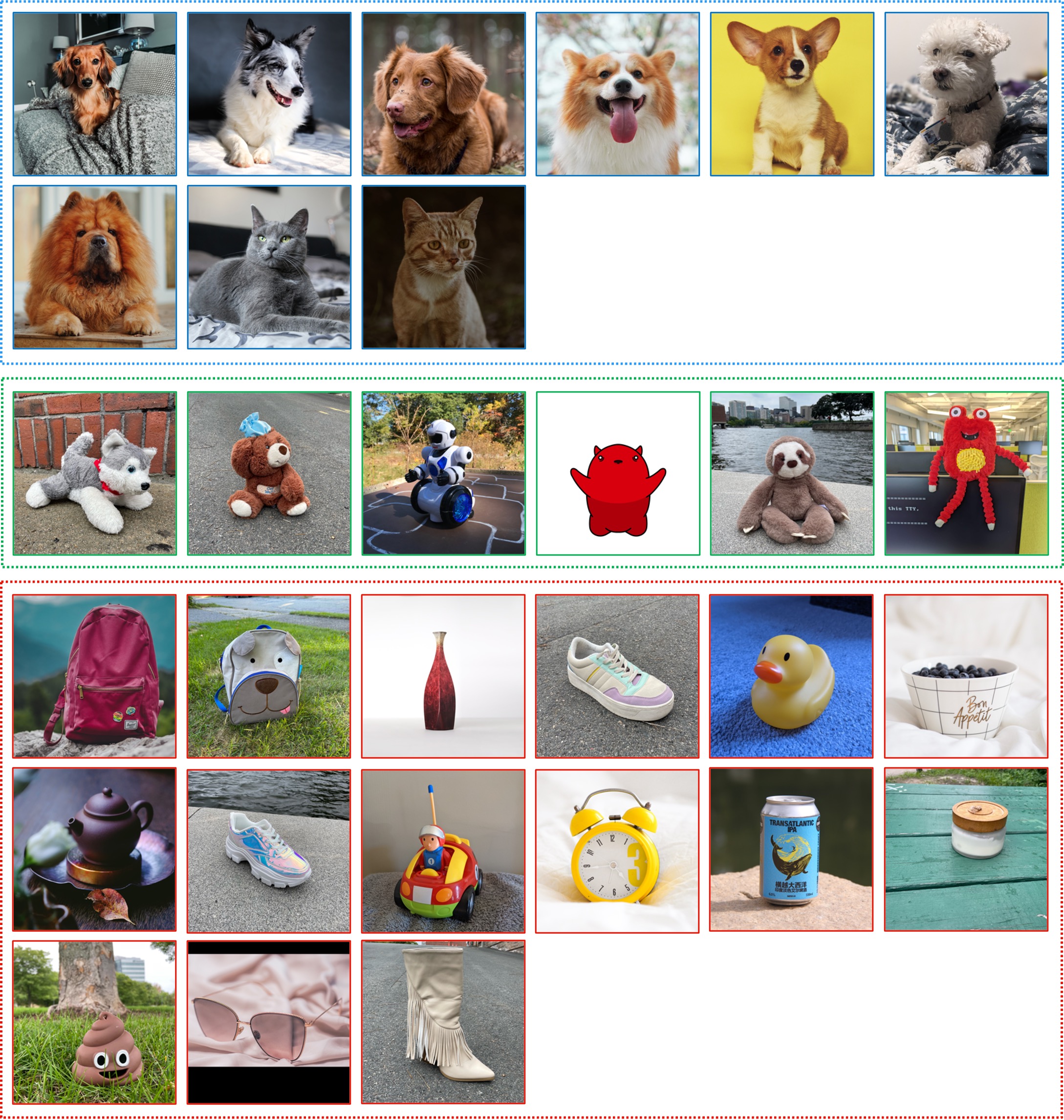}
    \vspace{-0.1in}
    \caption{Images used in quantitative evaluation.}
    \label{fig:selected_subject}
    \vspace{-0.1in}
\end{figure}

\paragraph{Subject images in quantitative evaluation}
The subject images used in our quantitative evaluation are presented in Figure \ref{fig:selected_subject}, where animals, active objects and static objects are indicated by blue, green and red color respectively. High-resolution images can be found from the official release of DreamBench at ~\url{https://github.com/google/dreambooth}.

\begin{figure}[t!]
    \centering
    \includegraphics[width=0.95\linewidth]{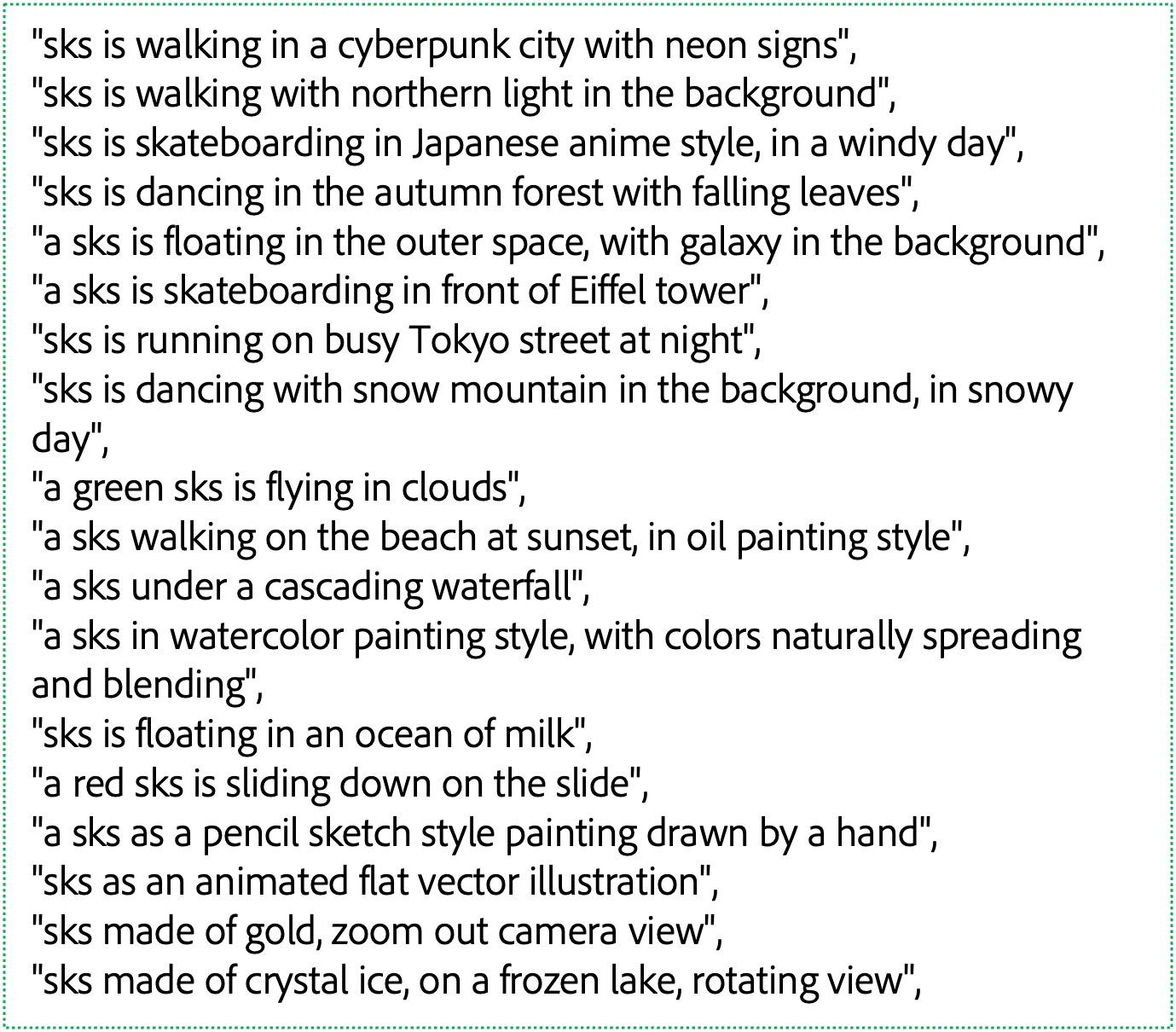}
    \vspace{-0.1in}
    \caption{Testing prompts designed for active objects.}
    \label{fig:active_object_prompt}
    \vspace{-0.1in}
\end{figure}

\begin{figure}[h!]
    \centering
    \includegraphics[width=0.95\linewidth]{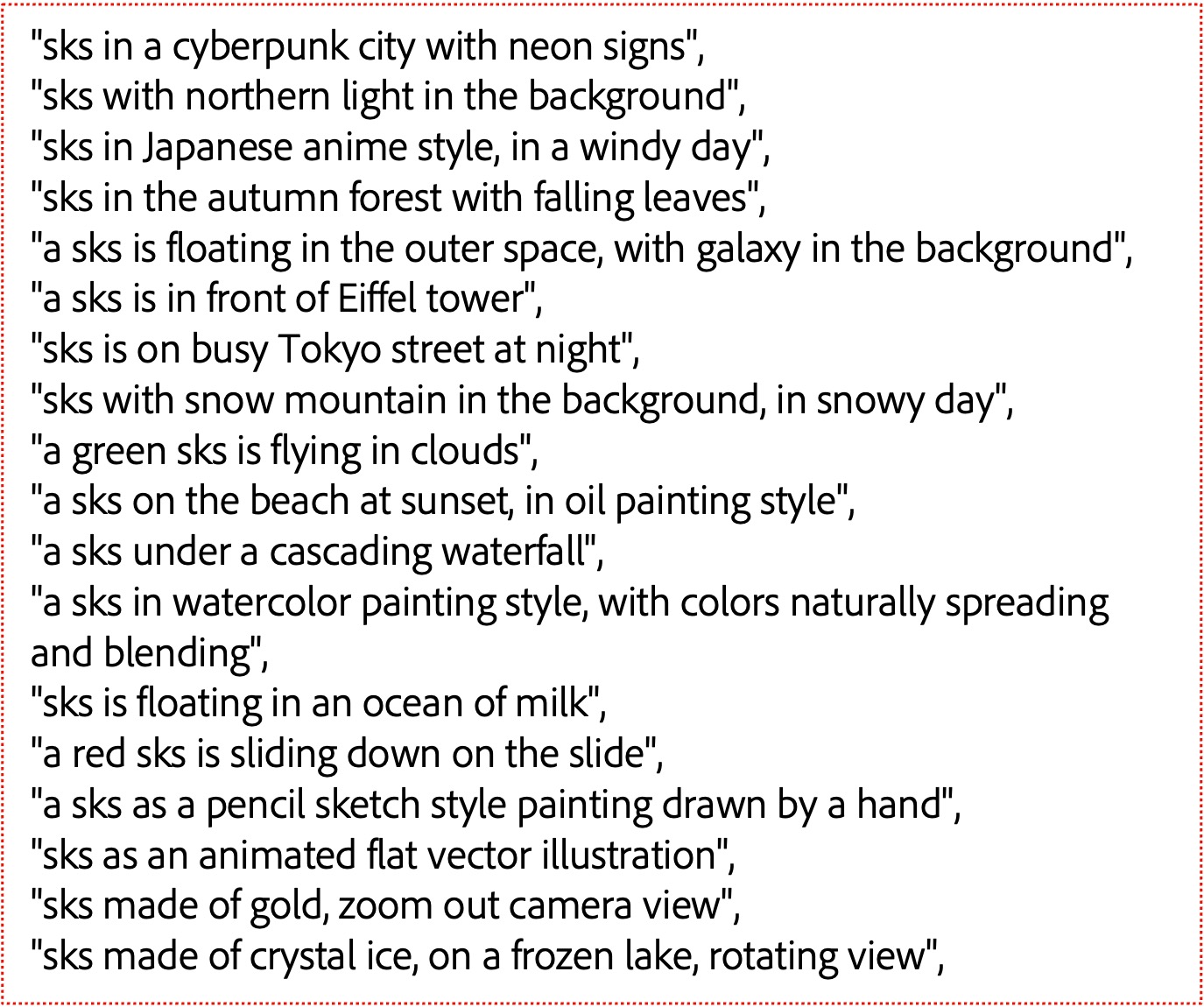}
    \vspace{-0.1in}
    \caption{Testing prompts designed for static objects.}
    \label{fig:static_object_prompt}
    \vspace{-0.1in}
\end{figure}

\begin{figure}[t!]
    \centering
    \includegraphics[width=0.95\linewidth]{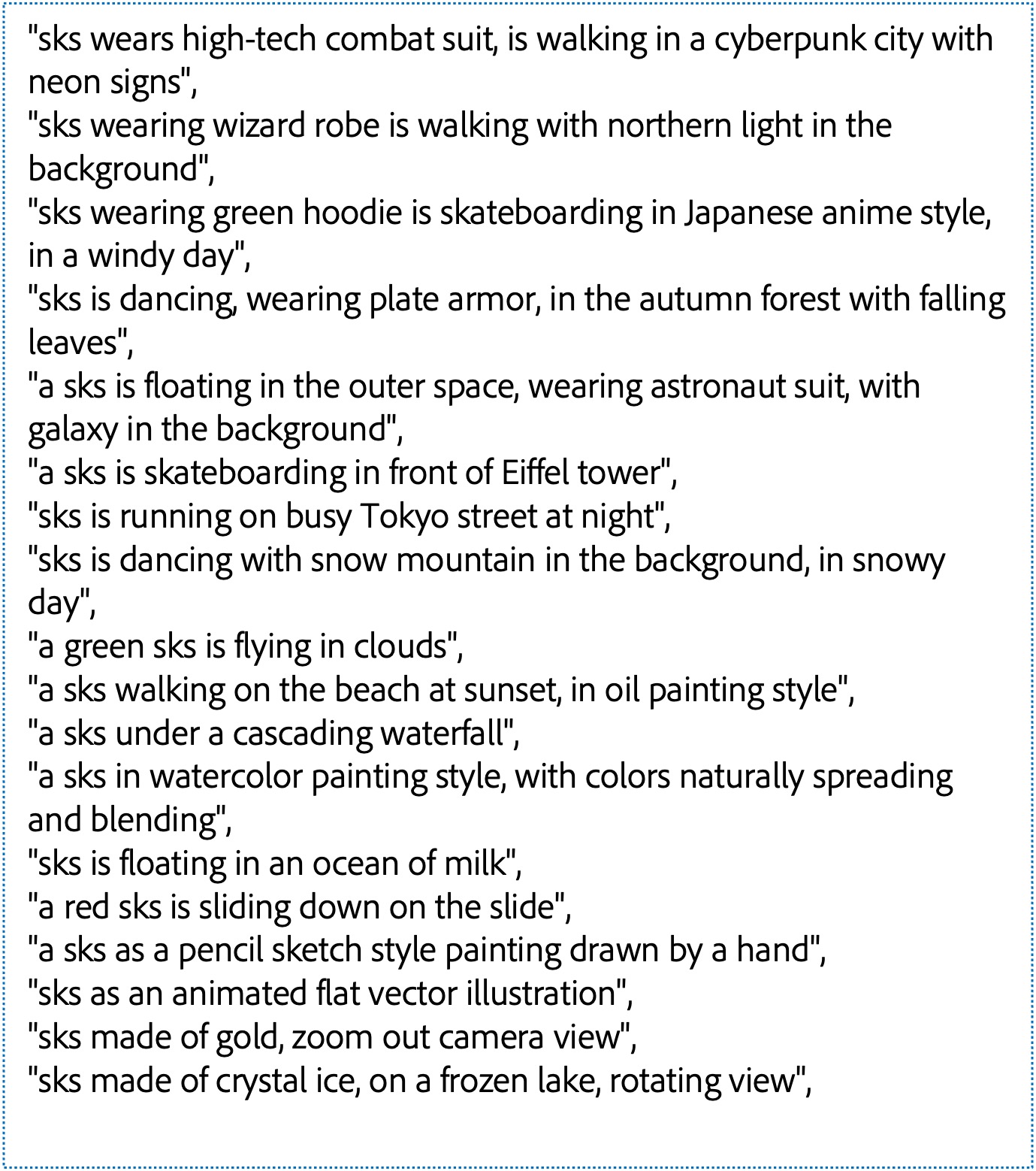}
    \vspace{-0.1in}
    \caption{Testing prompts designed for animals.}
    \label{fig:animal_prompt}
    \vspace{-0.1in}
\end{figure}

\paragraph{Designed testing prompts}
We provide our testing prompts for quantitative evaluation in Figure \ref{fig:animal_prompt}, Figure \ref{fig:active_object_prompt} and Figure \ref{fig:static_object_prompt}. We use ``sks" as a special token, which will be replaced by corresponding noun such as ``dog" at testing.

\end{document}